\documentclass[review,12pt]{elsarticle}

\usepackage[numbers]{natbib}

\usepackage{tikz}

\usetikzlibrary{arrows.meta}     
\usetikzlibrary{positioning}     
\usetikzlibrary{calc}            
\usetikzlibrary{shapes.geometric}

\usepackage{tabularx}
\usepackage{subcaption}
\usepackage{multirow}
\usepackage{placeins}
\usepackage{xcolor}
\usepackage{booktabs}
\usepackage{makecell}
\usepackage{siunitx}
\usepackage{caption}
\usepackage{longtable, array}
\usepackage{bm}
\usepackage{algorithm}
\usepackage{algpseudocode}
\usepackage{amsmath,amssymb}
\usepackage{tikz}
\usetikzlibrary{arrows.meta,positioning,fit,backgrounds,shapes.geometric,decorations.pathreplacing,calligraphy,decorations.pathmorphing,calc}
\usetikzlibrary{decorations.text,decorations.pathmorphing}

\usepackage{pifont}
\usepackage{setspace}
\usepackage[colorlinks=true,
            citecolor=blue,
            linkcolor=blue,
            urlcolor=blue]{hyperref}
\definecolor{inputcol}{RGB}{100,116,139}   
\definecolor{hiddencol}{RGB}{37,99,235}     
\definecolor{outcorrect}{RGB}{22,163,74}    
\definecolor{outother}{RGB}{148,163,184}    
\definecolor{ltpcol}{RGB}{22,163,74}        
\definecolor{ltdcol}{RGB}{220,38,38}        
\definecolor{stage1col}{RGB}{180,83,9}      
\definecolor{stage2col}{RGB}{109,40,217}    
\definecolor{panelbg}{RGB}{248,250,252}
\definecolor{panelline}{RGB}{203,213,225}

\makeatletter
\def\emailauthor#1#2{}
\makeatother

\begin{document}

\begin{frontmatter}


\title{Building Supervision into Hebbian Plasticity through Spike Agreement}


\author[inst1]{Gouri Lakshmi S}

\author[inst1]{Athira Chandrasekharan}

\author[inst1]{Harshit Kumar}

\author[inst1]{Muhammed Sahad E}

\author[inst1]{Bikas Das}

\author[inst1]{Saptarshi Bej\corref{cor1}}

\cortext[cor1]{Corresponding author}

\ead{sbej24@iisertvm.ac.in}


\address[inst1]{Indian Institute of Science Education and Research Thiruvananthapuram,
Vithura, Thiruvananthapuram, Kerala 695551, India}






\begin{abstract}
Supervised learning in spiking neural networks (SNNs) typically requires
either gradient-based backpropagation, which sacrifices the Hebbian,
spike-driven character of biological plasticity, or reward-modulated
Spike-Timing-Dependent Plasticity (STDP), in which class supervision enters
only as a scalar gate on an otherwise class-agnostic correlation signal.
We propose a novel Hebbian plasticity rule, \textit{Supervised Spike Agreement-Dependent Plasticity} (Supervised SADP), a gradient-free supervised learning algorithm in which class information is embedded directly into the Hebbian plasticity computation itself, rather than being introduced through reward modulation, teacher-forcing signals, or gradient-based error propagation.

The algorithm trains the output layer via a supervised Hebbian rule that
encodes class labels into output spike patterns, then trains the hidden layer
by measuring each hidden neuron's chance-corrected temporal agreement,
Cohen's~$\kappa$, with the correct-class output spike train produced by the
forward pass without any gradient computation or external reward signal.
A $K$-shift extension aggregates agreement over a window of temporal offsets,
providing robustness to spike-timing jitter at linear computational cost, and
an optional reward-modulation term introduces a global third plasticity factor.

We evaluate Supervised SADP against reward-modulated STDP across six datasets
spanning benchmark image classification (MNIST, Fashion-MNIST, CIFAR-10) and
medical imaging (colon histopathology, lung histopathology, brain MRI tumour),
four input encoding strategies, $K_{\text{shift}} \in \{5,25\}$, and three
reward modes (\texttt{none}, \texttt{binary}, \texttt{margin}).

Supervised SADP outperforms STDP in a significant majority of encoding-reward-mode
comparisons.
Under purely gradient-free Poisson encoding, SADP achieves 86.46\% on MNIST
and 76.62\% on Fashion-MNIST, outperforming the best STDP configurations
under the same network architecture by 23.66 and 23.29 percentage points,
respectively.

Across the encodings tested, including CNN-extracted features as a high-end
encoding baseline, SADP outperforms STDP in the large majority of cells and
trains $1.47\times$ faster on average, with up to $2.86\times$ speedup under
high-dimensional Poisson inputs.

These results position Supervised SADP as a stable, efficient, and
gradient-free alternative to reward-modulated STDP for supervised SNN
learning, with demonstrated applicability to benchmark vision and medical
imaging datasets.
\end{abstract}



\end{frontmatter}

\section{Introduction}
\label{sec:introduction}

Spiking neural networks (SNNs) are the biologically closest computational model of neural information processing, encoding information in the timing and rate of discrete spike events rather than continuous activations.
This design offers intrinsic advantages for energy-efficient neuromorphic hardware, where each synaptic event incurs a fixed, small energy cost, and for modelling learning in biological nervous systems, where synaptic plasticity is driven by local spike-time statistics rather than global error
signals~\cite{davies2018loihi, gerstner2014neuronal}.
The defining challenge in this field is \emph{supervised learning}: how can an SNN be trained to classify labelled inputs without gradient computation or error backpropagation, while retaining the Hebbian, spike-driven character of biological plasticity?

The dominant local learning rule for SNNs is Spike-Timing-Dependent Plasticity (STDP), which modifies synaptic weights according to the relative timing of pre- and post-synaptic spikes~\cite{bi1998synaptic, markram1997regulation}. STDP-based models have demonstrated effective unsupervised feature extraction in visual recognition tasks~\cite{MasquelierThorpe2007, Kheradpisheh2018,
Ferre2018, diehl2015unsupervised}.
To introduce supervision, reward-modulated STDP incorporates a global reinforcement signal as a third plasticity factor~\cite{izhikevich2007solving,
fremaux2016neuromodulated}, while teacher-forcing methods inject external currents to drive neurons toward desired firing patterns~\cite{Hao2020SymmetricSTDP}.
Gradient-based approaches, including surrogate gradient descent and e-prop, achieve strong supervised accuracy by approximating backpropagation through spike discontinuities~\cite{neftci2019surrogate,
bellec2020solution}, but they abandon strict locality and biological plausibility.
More recently, Spike Agreement-Dependent Plasticity (SADP) was introduced as a computationally efficient local alternative to STDP, replacing pairwise spike-timing comparisons with chance-corrected population agreement
statistics computed over extended temporal windows, yielding faster and more stable unsupervised learning~\cite{bej2025sadp}.

Two concrete gaps remain in the landscape of Hebbian supervised learning
for SNNs.
 
\textbf{Gap 1: class supervision is absent from the Hebbian correlation term
in all existing Hebbian SNN rules.}
The Hebbian principle states that a synapse from neuron $i$ to neuron $j$
is modified in proportion to the correlation between their activities.
In STDP and its reward-modulated variants, this correlation is the relative
timing of pre-synaptic and post-synaptic spikes, a quantity that is
computed without any reference to the correct class label.
Class information enters only as a scalar reward $r$ that gates or scales
the update \emph{after} the Hebbian correlation has been formed; the
direction of plasticity is set by timing coincidences, not by the target.
Teacher-forcing methods embed class information by injecting external
currents that directly impose the desired post-synaptic activity during
learning~\cite{Hao2020SymmetricSTDP}. Although the resulting weight updates
may remain local, the supervision is introduced by modifying the network's
forward dynamics rather than by evaluating activity generated naturally by
the network itself. The fundamental gap is therefore: \emph{no existing gradient-free Hebbian
rule for SNNs defines the plasticity signal itself by agreement with a
class-specific activity pattern that emerges from the network's own forward
dynamics.}
As a consequence, gradient-free Hebbian rules either require a carefully
engineered reward to inject any class information at all, and collapse to
near-random without it, or require external interference with the
post-synaptic activity.
 
\textbf{Gap 2: the agreement-based plasticity framework is unsupervised.}
Spike Agreement-Dependent Plasticity (SADP)~\cite{bej2025sadp} replaces
STDP's pairwise trace-based correlation with a chance-corrected agreement
statistic (Cohen's~$\kappa$) computed over the full simulation window,
yielding faster and more computationally efficient weight updates.
However, unsupervised SADP computes agreement between arbitrary pre- and
post-synaptic neurons without reference to class labels, limiting its
applicability to supervised classification tasks where the target class must
drive learning.

We address both gaps by introducing \emph{Supervised SADP}, a novel
learning algorithm that extends the SADP framework~\cite{bej2025sadp} to
labelled classification through a two-stage, gradient-free update rule.
The output layer is updated using a supervised Hebbian rule that encodes
the class label directly into the output spike pattern as a teaching signal.
The hidden layer is then updated by measuring the \emph{chance-corrected
spike agreement} (Cohen's $\kappa$) between each hidden neuron's spike train
and the correct-class output neuron's spike train without any gradient
computation.
The correct-class output spike train is produced by the forward pass itself
and serves as a cross-layer supervisory reference; no weight-specific error
signals are propagated.
A $K$-shifted extension aggregates agreement over a range of temporal offsets,
providing robustness to axonal delays and spike-timing jitter at negligible
additional cost.
An optional reward-modulation term adds a trial-level third factor, but
crucially, the algorithm is effective even without it: unlike STDP, the
agreement signal already encodes class supervision through the forward-produced
output spike train, requiring no external reward for learning to succeed.

Evaluated on six datasets spanning benchmark image classification (MNIST,
Fashion-MNIST, CIFAR-10) and medical imaging (colon histopathology, lung
histopathology, brain MRI), Supervised SADP outperforms
reward-modulated STDP in the large majority of dataset--encoding cells.
The advantage is most pronounced under encodings with weak discriminative
structure, where STDP without careful hyperparameter tuning collapses
while SADP continues to learn effectively.
Supervised SADP also trains faster on average, with the gap widest on
high-dimensional Poisson-encoded inputs where STDP's trace-maintenance
overhead is greatest.

These results establish Supervised SADP as a principled, gradient-free
alternative for supervised SNN learning with direct implications for
neuromorphic hardware, where layer-wise backward error propagation is architecturally
infeasible~\cite{davies2018loihi}.
The rule's compatibility with diverse input encodings and its robustness
to reward signal design make it applicable to resource-constrained settings
including medical image analysis, where data efficiency and interpretability
are important alongside accuracy.

\paragraph{Contributions.}
\begin{enumerate}
    \item \textbf{Supervised SADP learning rule.}
          A novel gradient-free, backpropagation-free supervised Hebbian
          learning rule for SNNs based on chance-corrected spike agreement
          (Cohen's $\kappa$) between hidden neurons and the correct-class
          output neuron's forward-produced spike train.
          Unlike STDP, the rule encodes class supervision without requiring
          any external reward signal. The principal contribution of this work
          is the plasticity rule itself; the network architecture is
          intentionally minimal to isolate and evaluate the behaviour of the
          proposed learning rule.

  \item \textbf{$K$-shifted temporal agreement.}
        An extension that aggregates $\kappa$ over a window of temporal
        offsets, providing robustness to axonal delays and spike-timing
        jitter in $\mathcal{O}(T)$ time per neuron.

  \item \textbf{Optional reward-modulated three-factor rule.}
        Binary and margin reward modes that optionally add a global
        neuromodulatory term, consistent with biological three-factor
        plasticity, as an independently switchable enhancement.

  \item \textbf{Comprehensive evaluation.}
        Systematic comparison against reward-modulated STDP across six
        datasets, four encoding strategies, and two hyperparameter settings
        per algorithm, including the first evaluation of agreement-based
        plasticity on medical imaging datasets.
\end{enumerate}

The remainder of the paper is organised as follows.
Section~\ref{sec:related_research} reviews relevant prior work on STDP,
supervised SNN learning, and population-level plasticity.
Section~\ref{sec:methodology} derives the Supervised SADP model and learning
rule in full.
Experimental details are given in Section~\ref{sec:Experiments}, and
results are presented and discussed in Section~\ref{sec:results}.

\section{Related Work}
\label{sec:related_research}

\subsection{Supervised Learning in Spiking Neural Networks}
\label{sec:rw_supervised}

Training SNNs for supervised tasks is fundamentally difficult because the
spike function is discontinuous, making direct gradient computation
undefined.
The dominant practical solution is \emph{surrogate gradient learning}, which
substitutes a smooth approximation for the spike-function derivative during
the backward pass~\cite{neftci2019surrogate}.
Surrogate methods have achieved strong accuracy on standard vision benchmarks
and have been extended to recurrent and multi-layer architectures; however,
they require a full backward pass through the network and thus cannot be
implemented on neuromorphic hardware that lacks backward connectivity.
A related approach, \emph{eligibility propagation} (e-prop), computes
approximate gradients using local eligibility traces that can, in principle,
be maintained on-chip~\cite{bellec2020solution}, yet it still requires
broadcasting a global error signal to each synapse, breaking strict
synaptic locality.

A second family of supervised methods uses \emph{teacher forcing}: external
currents drive postsynaptic neurons toward desired spike patterns, combined
with local Hebbian updates~\cite{Hao2020SymmetricSTDP, Zhao2020GLSNN}.
Symmetric STDP with teacher signals achieves competitive accuracy on
Fashion-MNIST but remains sensitive to the strength and duration of the
injected teaching current, and does not generalise straightforwardly to
arbitrary target label structures.
\emph{Reward-modulated} approaches add a global reward or prediction-error
signal as a third plasticity factor, bridging Hebbian learning and
reinforcement signals~\cite{izhikevich2007solving, fremaux2016neuromodulated, kusmierz2017learning}.
These rules are biologically interpretable and neuromorphic-compatible, but
they require careful joint selection of STDP time constants and reward
functions; under unfavourable combinations, performance collapses to
near-random, as we demonstrate empirically.

Supervised SADP occupies a distinct position: it is gradient-free, requires
no backward pass and no teacher-forcing current, and encodes class supervision
through a forward-computed agreement statistic between hidden neurons and the
correct-class output spike train, rather than through an external reward
or a layer-wise error signal.

\subsection{Classical STDP and Its Limitations for Supervised Tasks}
\label{sec:rw_stdp}

STDP modifies a synapse based on the timing difference $\Delta t =
t_{\text{post}} - t_{\text{pre}}$ between individual spike pairs, inducing
Long-Term Potentiation (LTP) when presynaptic activity precedes postsynaptic
firing and Long-Term Depression (LTD) in the reverse order~\cite{bi1998synaptic,
markram1997regulation}.
Extensions including triplet STDP~\cite{pfister2006triplets}, voltage-based
STDP~\cite{clopath2010connectivity}, and higher-order models have improved
biological fidelity and pattern-selectivity~\cite{clopath2010voltage,
morrison2008phenomenological}.
In unsupervised settings, STDP combined with competitive inhibition or
convolutional architectures has achieved notable results on digit recognition
and simple object classification~\cite{MasquelierThorpe2007, Kheradpisheh2018,
diehl2015unsupervised}.

Two structural limitations constrain STDP in supervised contexts.
\textbf{Sensitivity to spike timing.} STDP requires millisecond-scale
precision; small jitter in spike timing reverses the sign of the update,
making training unstable in noisy or large-scale circuits~\cite{Markram2011,
Markram2012}.
\textbf{Computational cost.} Each synaptic update requires evaluating every
pre-post spike pair, incurring per-synapse cost that grows with the square
of the spike count per timestep, expensive for high-firing-rate encodings
or large networks~\cite{Tian2025}.
Recent work on synchrony-dependent plasticity (SSDP, S2-STDP) addresses
timing sensitivity by aggregating over neural populations~\cite{Tian2025,
Chawla1999}, but these remain largely unsupervised and do not incorporate
class label information.

\subsection{Agreement-Based Plasticity and Unsupervised SADP}
\label{sec:rw_sadp}

Spike Agreement-Dependent Plasticity (SADP) was introduced by Bej et
al.~\cite{bej2025sadp} as a biologically inspired learning rule that replaces
pairwise spike-timing evaluation with a chance-corrected statistical measure
of agreement between spike trains computed over an entire temporal window.
Using Cohen's $\kappa$ as the agreement statistic~\cite{cohen1960coefficient},
SADP is insensitive to coincidental agreement arising from shared average
firing rates, a confound that affects raw correlation-based rules  and
reduces the per-synapse update complexity to $\mathcal{O}(T)$.
Empirically, unsupervised SADP demonstrates improved training stability
and faster convergence compared with STDP across several encoding strategies,
particularly under Poisson-encoded inputs where pairwise spike-timing
evaluation is most expensive~\cite{bej2025sadp}.

However, unsupervised SADP computes agreement without reference to class
labels: the agreement is measured between pre- and post-synaptic neurons
arbitrarily, without directing plasticity toward a classification objective.
The present work addresses this by introducing a supervised reference signal
for the agreement computation (the correct-class output neuron's spike
train, produced by the forward pass) and coupling it with a supervised
Hebbian output update, enabling end-to-end supervised classification
without any gradient computation or layer-wise error propagation.

\subsection{Benchmarking Against Prior STDP-Based Supervised Methods}
\label{sec:rw_benchmark}

Table~\ref{tab:prior_work} summarises reported classification accuracies
of supervised SNN learning rules on standard vision benchmarks.
The results are drawn from the literature and are reproduced for reference
to contextualise the performance regime that agreement-based methods must
match or exceed.
Reward-modulated and synchrony-based STDP methods achieve competitive
results when hyperparameters are carefully tuned, but exhibit high variance
across configurations and collapse under suboptimal settings.
Gradient-based surrogate methods (not shown) typically achieve higher
accuracy at the cost of backward connectivity, which is unavailable on
most neuromorphic processors.
Our Supervised SADP results are reported in full in
Section~\ref{sec:results}.

\begin{table}[htbp]
\centering
\caption{%
  Reported classification accuracy (\%) of SNN learning rules on standard
  vision benchmarks, compiled from the literature.
  All entries are prior work included for contextualisation.
  All results from~\cite{Tian2025} unless otherwise noted.
  BP-SNN denotes a backpropagation-based SNN baseline (result from~\cite{Tian2025}).%
}
\label{tab:prior_work}
\setlength{\tabcolsep}{6pt}
\begin{tabular}{@{}llc@{}}
\toprule
\textbf{Dataset} & \textbf{Method} & \textbf{Accuracy (\%)} \\
\midrule
\multirow{6}{*}{Fashion-MNIST}
 & Sym-STDP~\cite{Hao2020SymmetricSTDP}   & 85.3 \\
 & GLSNN~\cite{Zhao2020GLSNN}             & 89.1 \\
 & STB-STDP~\cite{Tian2025}               & 87.0 \\
 & R-STDP~\cite{Mozafari2018}             & 83.3 \\
 & SSTDP~\cite{Tian2025}                  & 85.2 \\
 & S2-STDP~\cite{Tian2025}                & 85.9 \\
\midrule
\multirow{4}{*}{N-MNIST}
 & STDP~\cite{diehl2015unsupervised}                   & 98.7 \\
 & Synaptic Kernel Inverse~\cite{Cohen2016SkimmingDigits} & 92.9 \\
 & BP-SNN                                              & 92.7 \\
 & STDP+CDNA-SNN~\cite{Saranirad2024CDNASNN}          & 98.4 \\
\midrule
\multirow{5}{*}{CIFAR-10}
 & STB-STDP~\cite{Tian2025}               & 32.9 \\
 & R-STDP~\cite{Deepr2024}                & 51.7 \\
 & S2-STDP~\cite{Tian2025}                & 61.8 \\
 & SSTDP~\cite{Tian2025}                  & 60.8 \\
 & NCG-S2-STDP~\cite{Goupy2024NCGSTDP}   & 66.4 \\
\bottomrule
\end{tabular}
\end{table}

\begin{table}[htbp]
\centering
\caption{Comparison of major supervised learning paradigms for supervised SNN learning.}
\label{tab:positioning}

\footnotesize
\begin{tabularx}{\columnwidth}{>{\raggedright\arraybackslash}p{3.0cm}
>{\centering\arraybackslash}p{1.8cm}
>{\raggedright\arraybackslash}X
>{\raggedright\arraybackslash}X}
\toprule
\textbf{Method} &
\textbf{Gradient-free} &
\textbf{Local plasticity} &
\textbf{Biological plausibility} \\
\midrule
Surrogate Gradient/BPTT
& No
& Weight-specific error gradients
& Low \\

e-prop
& Partial
& Eligibility traces + broadcast error
& Moderate \\

Teacher Forcing
& Yes
& Local updates with externally imposed activity
& Moderate \\

Reward-modulated STDP
& Yes
& Local Hebbian + global reward
& High \\

\textbf{Supervised SADP}
& Yes
& Local Hebbian + forward-generated supervisory reference
& High \\
\bottomrule
\end{tabularx}
\end{table}

Table~\ref{tab:positioning} conceptually positions the major supervised
learning paradigms for spiking neural networks. While teacher-forcing
approaches are also gradient-free, they rely on externally imposed
postsynaptic activity during learning, thereby altering the network's
natural forward dynamics. In contrast, both reward-modulated STDP and
Supervised SADP preserve forward network dynamics and update synapses
through local Hebbian mechanisms without backward error propagation.
Both methods therefore belong to the same class of gradient-free Hebbian
learning rules, differing primarily in how class supervision enters the
plasticity rule. Reward-modulated STDP introduces supervision through a
delayed global reward signal after the Hebbian correlation has been
computed, whereas Supervised SADP embeds class information directly into
the plasticity signal by measuring agreement with the correct-class
output activity generated by the network's own forward dynamics.
Consequently, reward-modulated STDP represents the closest existing
methodological baseline for evaluating the proposed learning rule.

\section{Methodology}
\label{sec:methodology}

We propose \textit{Supervised Spike Agreement-Dependent Plasticity} (SADP),
a gradient-free Hebbian learning algorithm for spiking neural networks
(SNNs).
SADP belongs to the family of Hebbian learning rules: weight updates require
no gradient computation and no backward pass.
The output layer update uses a local target-error signal at the output
neurons; this error does not propagate backward to the hidden layer.
The hidden layer update is driven by a \emph{chance-corrected spike agreement}
signal, Cohen's $\kappa$ computed between a hidden neuron's spike train and
the spike train of the correct-class output neuron produced by the forward
pass, which plays the role that eligibility traces play in STDP, but
provides a more informative, calibrated measure of temporal correlation
without requiring per-timestep trace maintenance.
The class label enters the hidden layer update through this forward-produced
reference signal, not through any weight-specific error propagation.
The sections below derive the network model, the two-stage update rule,
and two optional extensions (temporal shift tolerance and reward modulation)
that address stability and generalisation.

Throughout the derivation, the superscripts $h$ and $o$ denote hidden and
output quantities respectively; subscripts $i$, $j$, $k$ index input, hidden,
and output neurons; $t$ denotes timestep; and $y \in \{1,\ldots,N_c\}$ denotes
the true class label.
Bold symbols denote matrices or vectors; plain symbols denote scalars or
individual elements.
All derivations are stated for a single sample; the batch extension (averaging
over a mini-batch of size $B$) is straightforward and is used in practice.

\subsection{Network Architecture and Neuron Model}
\label{sec:network}
 
The contribution of this work is the proposed Hebbian plasticity rule rather than the neural network architecture. We therefore intentionally employ a simple three-layer feedforward SNN so that any performance improvements can be attributed to the learning rule itself rather than architectural design choices. The proposed learning rule is independent of this architecture and can be incorporated into other feedforward SNNs.
As illustrated in Figure~\ref{fig:sadp_overview}, the network consists of
three layers: an input layer with $N_i$ neurons,
a single hidden layer with $N_h$ neurons, and an output layer with $N_c$
neurons (one per class).
Connections from input to hidden are parameterised by
$\bm{W}_1 \in \mathbb{R}^{N_i \times N_h}$, and from hidden to output by
$\bm{W}_2 \in \mathbb{R}^{N_h \times N_c}$.
 
Each neuron follows a discrete-time \textit{Leaky Integrate-and-Fire} (LIF)
model~\cite{mainen1995reliability} simulated over $T$ timesteps.
For a hidden neuron $j$, the membrane potential $V^h_j(t)$ evolves as
\begin{equation}
  V^h_j(t) \;=\; \lambda\,V^h_j(t-1) \;+\; \sum_{i=1}^{N_i} W_{ij}^{1}\,s_i(t),
  \label{eq:lif_hidden}
\end{equation}
where $\lambda \in (0,1)$ is the leak (membrane decay) constant,
$s_i(t) \in \{0,1\}$ is the spike emitted by input neuron $i$ at timestep
$t$, and $W_{ij}^1$ is the corresponding synaptic weight.
Neuron $j$ fires when $V^h_j(t)$ exceeds a threshold $\theta_h$:
\begin{equation}
  s^h_j(t) \;=\; \mathbf{1}\!\left[V^h_j(t) > \theta_h\right],
  \label{eq:spike_hidden}
\end{equation}
and the membrane potential is immediately reset to zero upon firing:
$V^h_j(t) \leftarrow 0$ if $s^h_j(t)=1$.
Output neurons follow the same dynamics with weights $\bm{W}_2$ and a
potentially different threshold $\theta_o$:
\begin{equation}
  V^o_k(t) \;=\; \lambda\,V^o_k(t-1) \;+\; \sum_{j=1}^{N_h} W_{jk}^{2}\,s^h_j(t),
  \qquad
  s^o_k(t) \;=\; \mathbf{1}\!\left[V^o_k(t) > \theta_o\right].
  \label{eq:lif_output}
\end{equation}
 
The predicted class $\hat{y}$ is determined by the \emph{spike-count} rule:
the output neuron that accumulates the greatest number of spikes over the
entire simulation window is selected,
\begin{equation}
  \hat{y} \;=\; \operatorname*{arg\,max}_{k} \sum_{t=1}^{T} s^o_k(t).
  \label{eq:readout}
\end{equation}
This rate-coding readout requires no additional trained components.

\begin{figure}[!htbp]
\centering
\resizebox{\linewidth}{!}{%
\begin{tikzpicture}[
  >=Latex,
  neuron/.style={circle, draw, minimum size=6.5mm, inner sep=0pt, line width=0.7pt},
  fwd/.style={-{Latex[length=1.6mm]}, draw=black!35, line width=0.5pt},
  lyr/.style={font=\small\bfseries},
  eq/.style={font=\footnotesize},
  stagebox/.style={rounded corners=3pt, draw, fill=panelbg, line width=1pt, align=left, inner sep=6pt, text width=6.0cm},
  every node/.style={font=\sffamily}
]
\def\xin{0}\def\xhid{3.4}\def\xout{6.8}
\foreach \y in {0,1,2,3}{ \node[neuron, fill=inputcol!25, draw=inputcol] (i\y) at (\xin,\y) {}; }
\foreach \y in {0,1,2,3}{ \node[neuron, fill=hiddencol!22, draw=hiddencol] (h\y) at (\xhid,\y) {}; }
\node[neuron, fill=outcorrect!30, draw=outcorrect, line width=1.1pt] (o2) at (\xout,2.6) {};
\node[neuron, fill=outother!35, draw=outother] (o1) at (\xout,1.4) {};
\node[neuron, fill=outother!35, draw=outother] (o0) at (\xout,0.2) {};
\foreach \a in {0,1,2,3}{ \foreach \b in {0,1,2,3}{ \draw[fwd] (i\a) -- (h\b); }}
\foreach \a in {0,1,2,3}{ \foreach \b in {0,1,2}{ \draw[fwd] (h\a) -- (o\b); }}
\node[eq, hiddencol] at (1.7,3.9) {$\bm{W}_1$};
\node[eq, outcorrect!60!black] at (5.1,3.9) {$\bm{W}_2$};
\node[lyr, inputcol, align=center] at (\xin,-0.95) {Input layer\\[-1pt]{\scriptsize$\bm{x}\!\in\![0,1]^{N_i}\!\to s_i(t)$}};
\node[lyr, hiddencol, align=center] at (\xhid,-0.95) {Hidden layer\\[-1pt]{\scriptsize$s^h_j(t),\; N_h{=}256$}};
\node[lyr, outcorrect!60!black, align=center] at (\xout,-0.95) {Output layer\\[-1pt]{\scriptsize$s^o_k(t)$, one per class}};
\node[eq, outcorrect!60!black, anchor=west] at ($(o2)+(0.5,0)$) {\scriptsize correct class $y$};
\node[stagebox, draw=stage1col] (s1) at (2.2,-3.35) {%
  {\color{stage1col}\bfseries Stage 1 --- $\bm{W}_2$ update}\\[2pt]
  target $\tau_k(t)=\mathbf{1}[k{=}y]$\\
  error $e_k(t)=\tau_k(t)-s^o_k(t)$\\
  $\Delta W^2_{jk}=\eta_2\sum_t s^h_j(t)\,e_k(t)$};
\node[stagebox, draw=stage2col] (s2) at (9.2,-3.35) {%
  {\color{stage2col}\bfseries Stage 2 --- $\bm{W}_1$ update}\\[2pt]
  reference $s^{o,\star}(t)\equiv s^o_y(t)$\\
  $\kappa_j=\dfrac{p^{\mathrm{obs}}_j-p^{e}_j}{1-p^{e}_j}$\\[2pt]
  $\Delta W^1_{ij}=\eta_1\,\bar x_i\,\tilde\kappa_j$};
\draw[-{Latex[length=2.4mm]}, stage1col, line width=1.1pt] ($(o1)+(0,-0.3)$) to[out=-90,in=90] (s1.north);
\draw[-{Latex[length=2.4mm]}, stage2col, line width=1.1pt, dashed] (o2.east) to[out=0,in=90] (s2.north);
\end{tikzpicture}%
}
\caption{%
Overview of the proposed Supervised Spike Agreement-Dependent Plasticity
(Supervised SADP) framework. Forward propagation runs through the input,
hidden, and output layers; learning then proceeds in two gradient-free
stages. \emph{Stage~1} updates the output synapses $\bm{W}_2$ with a
supervised Hebbian rule driven by the class-encoded target spike train.
\emph{Stage~2} updates the hidden synapses $\bm{W}_1$ using the
chance-corrected spike agreement (Cohen's $\kappa$) between each hidden
neuron and the forward-generated spike train of the correct-class output
neuron. The architecture is intentionally minimal so that performance can be
attributed to the learning rule rather than to network design.}
\label{fig:sadp_overview}
\end{figure}

\subsection{The SADP Learning Rule}
\label{sec:sadp_rule}

Training proceeds by alternating a forward pass (Equations~\ref{eq:lif_hidden}%
--\ref{eq:lif_output}) with a two-stage weight update:
the output layer $\bm{W}_2$ is updated first using an explicit supervision
signal; the hidden layer $\bm{W}_1$ is then updated using the
\textit{spike agreement} between each hidden neuron and the now-updated
output layer's activity, without any gradient propagation.
 
\subsubsection{Stage 1: Output Layer Update}
\label{sec:w2_update}
 
The output layer is trained with a temporally-resolved supervised Hebbian
rule.
For a sample with true label $y$, define the binary target spike train for
output neuron $k$ as
\begin{equation}
  \tau_k(t) \;=\; \mathbf{1}[k = y] \;\in\; \{0,1\},
  \label{eq:target}
\end{equation}
i.e., the correct-class neuron is asked to fire at every timestep, and all
others to be silent.
The error at each timestep is $e_k(t) = \tau_k(t) - s^o_k(t)$.
The weight update is
\begin{equation}
  \Delta W^2_{jk}
  \;=\; \eta_2 \sum_{t=1}^{T} s^h_j(t)\,e_k(t)
  \;=\; \eta_2 \sum_{t=1}^{T} s^h_j(t)\,\bigl[\tau_k(t) - s^o_k(t)\bigr],
  \label{eq:w2_update}
\end{equation}
where $\eta_2 > 0$ is the output learning rate.
This rule potentiates synapses from hidden neurons that fire when the
output neuron should fire ($\tau_k=1$) and depresses those active when it
fires incorrectly ($s^o_k=1$ but $\tau_k=0$).
No target spike train is constructed for the hidden layer; the teaching
signal for $\bm{W}_1$ is derived purely from the network's own output
activity, as described next.
 
\subsubsection{Stage 2: Spike Agreement and Cohen's $\kappa$}
\label{sec:kappa}
 
The central object in SADP is the \emph{spike agreement} between a hidden
neuron and the \emph{correct-class output neuron} over the simulation window.
Let $s^{o,\star}(t) \equiv s^o_y(t)$ denote the spike train of the
correct-class output neuron.
 
For hidden neuron $j$, define the following quantities over $T$ timesteps:
\begin{align}
  p^h_j    &\;=\; \frac{1}{T}\sum_{t=1}^T s^h_j(t),  \label{eq:p_h}\\
  p^\star  &\;=\; \frac{1}{T}\sum_{t=1}^T s^{o,\star}(t), \label{eq:p_star}\\
  p^{\text{obs}}_j &\;=\; \frac{1}{T}\sum_{t=1}^T \mathbf{1}\!\left[s^h_j(t)=s^{o,\star}(t)\right]. \label{eq:p_obs}
\end{align}
Here $p^h_j$ is the empirical firing rate of hidden neuron $j$, $p^\star$ is
the firing rate of the correct-class output neuron, and $p^{\text{obs}}_j$ is
the observed fraction of timesteps on which neuron $j$ and the correct-class
output neuron are in agreement (both firing or both silent).
 
The expected agreement under the null hypothesis of \emph{independent} spike
trains with the same marginal rates is
\begin{equation}
  p^e_j \;=\; p^h_j\,p^\star \;+\; (1-p^h_j)(1-p^\star).
  \label{eq:p_e}
\end{equation}
Cohen's $\kappa$ for hidden neuron $j$ is defined as the chance-corrected
agreement:
\begin{equation}
  \kappa_j \;=\; \frac{p^{\text{obs}}_j - p^e_j}{1 - p^e_j}.
  \label{eq:kappa}
\end{equation}
$\kappa_j \in (-1, 1]$: a value of $0$ indicates agreement no better than
chance, positive values indicate positive correlation with the
correct-class output, and negative values indicate anti-correlation.
The chance correction is essential: a neuron that is silent throughout
($p^h_j=0$) trivially agrees with silent output timesteps;
raw agreement would reward such a neuron when the output fires rarely,
whereas $\kappa$ correctly assigns it a value of $0$.
 
\subsubsection{Stage 2 (continued): Hidden Layer Update}
\label{sec:w1_update}
 
The SADP update for $\bm{W}_1$ is a Hebbian rule in which $\kappa_j$ plays
the role of a \emph{post-synaptic modulator}: synapses onto hidden neuron
$j$ are strengthened when $\kappa_j > 0$ (the neuron co-fires with the
correct output) and weakened when $\kappa_j < 0$ (the neuron fires in
opposition to it):
\begin{equation}
  \Delta W^1_{ij}
  \;=\; \eta_1\,\bar{x}_i\,\kappa_j,
  \label{eq:w1_update}
\end{equation}
where $\bar{x}_i$ is the mean activity of input neuron $i$ over the
simulation window and $\eta_1 > 0$ is the input learning rate.
This is an outer-product Hebbian update: input $\times$ agreement.
The weight change is positive (LTP) when the input is active and the hidden
neuron co-fires with the correct output, and negative (LTD) otherwise.
 
There is no backpropagated gradient: the direction and magnitude of the
update are determined entirely by (i)~the pre-synaptic activity $\bar{x}_i$
and (ii)~the temporal agreement of the post-synaptic neuron with the
correct-class output's own spike train.
The output spike train therefore acts as a \emph{teaching pattern} for the
hidden layer, but this teaching is delivered through a forward measurement
of agreement rather than a backward computation of partial derivatives.

\subsection{Temporal Shift Tolerance via $K$-Shifted Agreement}
\label{sec:kshift}
 
In a physical SNN, synaptic transmission between the hidden and output layers
incurs an axonal delay.
Measuring agreement at exactly the same timestep ($d=0$) may therefore
underestimate the true functional relationship between a hidden neuron's
spikes and the consequent output spikes.
We generalise Equation~\eqref{eq:kappa} to allow a temporal displacement
$d \in \{-K,\ldots,+K\}$ between the two spike trains.
 
For a fixed offset $d$, let $T_d = T - |d|$ denote the length of the
overlapping portion.
The shifted hidden and output segments are
\begin{equation}
  s^h_j[d](t) \;=\; s^h_j(t), \quad
  s^{o,\star}[d](t) \;=\; s^{o,\star}(t+d),
  \quad t = 1,\ldots,T_d,
  \label{eq:shifted_segments}
\end{equation}
and the shifted kappa $\kappa_j(d)$ is computed from Equations
\eqref{eq:p_h}--\eqref{eq:kappa} applied to these segments.
The aggregate $K$-shifted agreement is the mean over all offsets:
\begin{equation}
  \tilde{\kappa}_j
  \;=\; \frac{1}{2K+1}\sum_{d=-K}^{K} \kappa_j(d).
  \label{eq:kappa_shifted}
\end{equation}
The hidden layer update (Equation~\ref{eq:w1_update}) replaces $\kappa_j$
with $\tilde{\kappa}_j$.
Setting $K=0$ recovers the original synchronous rule exactly.
Since $K$ is a fixed constant independent of dataset size, the computational
cost of the update remains $\mathcal{O}\!\left((2K+1)\,T\right) = \mathcal{O}(T)$
per hidden neuron, preserving the linear-time complexity of the base rule.
 
The $K$-shift serves a dual purpose: it provides robustness to small
jitter in spike timing, and it captures the case where a hidden neuron's
activity reliably \emph{precedes} (or follows) the output spike, which can
arise from the network's internal dynamics even in the absence of literal
axonal delay.
Figure~\ref{fig:kappa_kshift} illustrates the chance-corrected agreement
statistic and the $K$-shift aggregation mechanism.

\begin{figure}[!htbp]
\centering
\resizebox{\linewidth}{!}{%
\begin{tikzpicture}[
  >=Latex,
  lbl/.style={font=\footnotesize},
  tiny lbl/.style={font=\scriptsize},
  panel/.style={rounded corners=3pt, draw=panelline, fill=panelbg, line width=0.8pt},
  every node/.style={font=\sffamily}
]
\def\traceL{0}\def\traceR{11.6}\def\rowHid{2.0}\def\rowOut{0.6}
\node[tiny lbl, hiddencol, anchor=east] at (\traceL-0.35,\rowHid) {$s^h_j(t)$};
\node[tiny lbl, outcorrect!70!black, anchor=east] at (\traceL-0.35,\rowOut) {$s^{o,\star}(t)$};
\draw[draw=black!40, line width=0.6pt] (\traceL,\rowHid-0.35) -- (\traceR,\rowHid-0.35);
\draw[draw=black!40, line width=0.6pt] (\traceL,\rowOut-0.35) -- (\traceR,\rowOut-0.35);
\node[tiny lbl, black!45, anchor=west] at (\traceR+0.15,\rowHid-0.35) {$t$};
\node[tiny lbl, black!45, anchor=west] at (\traceR+0.15,\rowOut-0.35) {$t$};
\def\hidspikes{0,0,1,0,1,1,0,0,1,0,0,1,0,1,0,0,1,0,0,1,0,1,0,0,1}
\def\outspikes{0,0,0,1,0,1,1,0,0,1,0,0,1,0,1,0,0,1,0,0,1,0,1,0,0}
\pgfmathsetmacro{\nspk}{25}
\pgfmathsetmacro{\dx}{(\traceR-\traceL)/\nspk}
\foreach \v [count=\n from 0] in \hidspikes {
  \pgfmathsetmacro{\xpos}{\traceL + \n*\dx + \dx/2}
  \ifnum\v=1 \draw[draw=hiddencol, line width=1.6pt] (\xpos,\rowHid-0.32) -- (\xpos,\rowHid+0.32); \fi }
\foreach \v [count=\n from 0] in \outspikes {
  \pgfmathsetmacro{\xpos}{\traceL + \n*\dx + \dx/2}
  \ifnum\v=1 \draw[draw=outcorrect, line width=1.6pt] (\xpos,\rowOut-0.32) -- (\xpos,\rowOut+0.32); \fi }
\pgfmathsetmacro{\xmid}{1.55}
\draw[<->, dashed, stage2col, line width=1pt, yshift=-35pt] (\xmid-1.3,1.30) -- (\xmid+1.3,1.30);
\node[tiny lbl, stage2col, anchor=south, yshift=-30pt] at (\xmid,0.58) {$d\in\{-K,\ldots,+K\}$};
\node[tiny lbl, stage2col, align=center, anchor=north, yshift=-39pt] at (\xmid,0.98)
   {Shift output spike train\\Compute $\kappa_j(d)$};
\node[tiny lbl, align=left, anchor=north west, draw=stage2col, fill=white, rounded corners=2pt,
      text width=5.5cm, line width=0.8pt, inner sep=5pt] (kformula) at (\traceL-0.3,-1.55)
      {$p^{\text{obs}}_j(d)=\dfrac{1}{T_d}\displaystyle\sum_t \mathbf{1}[s^h_j(t){=}s^{o,\star}(t{+}d)]$\\[4pt]
       $\tilde\kappa_j=\dfrac{1}{2K+1}\displaystyle\sum_{d=-K}^{K}\kappa_j(d)$\\[2pt]
       robust to spike-timing jitter and axonal delay};
\node[tiny lbl, align=left, anchor=north west, draw=panelline, fill=white, rounded corners=2pt,
      text width=5.0cm, line width=0.7pt, inner sep=5pt] (legendB) at (6.1,-1.55)
   {\textcolor{hiddencol}{\rule{2.2mm}{0.9mm}} hidden neuron $j$ spikes\\[2pt]
    \textcolor{outcorrect}{\rule{2.2mm}{0.9mm}} correct-class output spikes\\[4pt]
    high overlap in window $\Rightarrow \tilde\kappa_j\!\uparrow \Rightarrow$ \textcolor{ltpcol}{\textbf{LTP}} on $W^1_{ij}$\\[2pt]
    low overlap / anti-corr. $\Rightarrow \tilde\kappa_j\!\downarrow \Rightarrow$ \textcolor{ltdcol}{\textbf{LTD}} on $W^1_{ij}$};
\begin{scope}[on background layer]
\coordinate (panelBsw) at ($(\traceL,3.0)+(-1.6,0)$);
\coordinate (panelBne) at ($(\traceR,-1.55)+(0.3,0)$);
\node[panel, fit=(panelBsw)(panelBne)(kformula)(legendB), inner sep=8pt] {};
\end{scope}
\end{tikzpicture}%
}
\caption{%
Chance-corrected spike agreement ($\kappa_j$) and the $K$-shift window.
The hidden neuron's spike train $s^h_j(t)$ is compared against the
correct-class output spike train $s^{o,\star}(t)$. For each temporal offset
$d\in\{-K,\ldots,+K\}$ the output train is shifted and the observed
agreement $p^{\mathrm{obs}}_j(d)$ is chance-corrected into $\kappa_j(d)$; the
offsets are averaged into $\tilde\kappa_j$ (Eq.~\ref{eq:kappa_shifted}). High
windowed overlap yields $\tilde\kappa_j\!>\!0$ and potentiates $W^1_{ij}$,
whereas anti-correlation yields $\tilde\kappa_j\!<\!0$ and depresses it,
giving robustness to spike-timing jitter and axonal delay at
$\mathcal{O}(T)$ cost.}
\label{fig:kappa_kshift}
\end{figure}

\subsection{Reward-Modulated Supervision}
\label{sec:reward}
 
The two-stage rule described above is a two-factor update: the weight change
depends on (i)~pre-synaptic activity and (ii)~the spike agreement of the
post-synaptic neuron with the correct-class output neuron.
A third, \emph{global} factor can be incorporated by scaling the hidden
layer update by a per-sample reward signal $r \in \mathbb{R}$:
\begin{equation}
  \Delta W^1_{ij} \;=\; \eta_1\,\bar{x}_i\,\tilde{\kappa}_j\,r.
  \label{eq:w1_reward}
\end{equation}
This three-factor rule is consistent with the neuromodulatory interpretation
of reward-modulated plasticity~\cite{fremaux2016neuromodulated}, where a
global signal (e.g., dopamine) gates synaptic modification.
No additional forward pass or computation beyond what the output layer
already produces is required to compute $r$.
 
We implement two reward functions:
\begin{itemize}
  \item \textbf{Binary reward.}
    $r = +1$ if the prediction is correct ($\hat{y} = y$), and $r = -1$
    otherwise.  This produces a simple sign flip of the update on error trials.
  \item \textbf{Margin reward.}
    Let $n_k = \sum_{t} s^o_k(t)$ denote the spike count of output
    neuron $k$.  The margin reward is
    \begin{equation}
      r \;=\; \frac{n_y - \max_{k \neq y} n_k}{T},
      \label{eq:margin_reward}
    \end{equation}
    i.e., the normalised difference between the spike count of the correct
    class and the highest-scoring competing class.
    This provides a continuous, confidence-calibrated signal: a large positive
    $r$ indicates a confident correct prediction, $r \approx 0$ indicates
    near-tie, and negative $r$ indicates an incorrect prediction.
\end{itemize}
Setting $r = 1$ for all samples (the \texttt{none} mode) reduces
Equation~\eqref{eq:w1_reward} back to the base rule \eqref{eq:w1_update}.
Importantly, even without reward modulation SADP does not require an
explicit reward: the spike agreement $\tilde{\kappa}_j$ already encodes
supervision through the correct-class output spike train, unlike
unsupervised STDP which has no access to label information.

\subsection{Weight Regularisation}
\label{sec:regularisation}

To prevent weight blow-up and maintain stable training dynamics,
two lightweight regularisation operations are applied after each mini-batch.
 
\paragraph{Input-to-hidden weights $\bm{W}_1$.}
Each column $\mathbf{w}^1_j$ (the weight vector projecting onto hidden
neuron $j$) is $\ell_2$-normalised after the update:
\begin{equation}
  \mathbf{w}^1_j \;\leftarrow\; \frac{\mathbf{w}^1_j}{\|\mathbf{w}^1_j\|_2 + \varepsilon},
  \label{eq:w1_norm}
\end{equation}
where $\varepsilon > 0$ is a small constant for numerical stability.
This constrains the effective weight scale to be uniform across hidden neurons
and is consistent with the biological notion of synaptic normalisation.
 
\paragraph{Hidden-to-output weights $\bm{W}_2$.}
Rather than normalising $\bm{W}_2$ (which would interfere with the
supervised Hebbian update), a weight decay with decay factor
$\gamma \in (0,1)$ is applied:
$\bm{W}_2 \leftarrow \gamma\,\bm{W}_2$,
followed by element-wise clipping to $[-c, c]$ to prevent runaway
potentiation of individual output connections.
Both $\bm{W}_1$ and $\bm{W}_2$ receive the multiplicative decay before
their respective normalisation or clipping steps.

\subsection{The SADP Algorithm}
\label{sec:algorithm}

Algorithm~\ref{alg:sadp} presents the complete SADP training procedure for
one mini-batch.
For clarity the pseudocode uses the single-sample convention; in practice
all operations are vectorised over a batch of size $B$.
 
\begin{algorithm}[htbp]
\begin{spacing}{1.0}
\footnotesize
\caption{Supervised SADP with one mini-batch update}
\label{alg:sadp}
\begin{algorithmic}[1]
\Require Feature vector $\mathbf{x} \in [0,1]^{N_i}$, label $y$,
         weights $\bm{W}_1, \bm{W}_2$,
         hyperparameters $T, \lambda, \theta_h, \theta_o, K, \eta_1, \eta_2, \gamma, c$,
         reward mode $\in \{\texttt{none}, \texttt{binary}, \texttt{margin}\}$
 
\Statex \textbf{--- Forward pass ---}
\State Encode $\mathbf{x}$ into spike train $s_i(t)$ for $t=1,\ldots,T$
       \Comment{encoding method described in Sec.~\ref{sec:encodings}}
\State Initialise $V^h_j(0) \leftarrow 0$ for all $j$;
       $V^o_k(0) \leftarrow 0$ for all $k$
 
\For{$t = 1$ \textbf{to} $T$}
  \State $V^h_j(t) \leftarrow \lambda\,V^h_j(t{-}1) + \sum_i W^1_{ij}\,s_i(t)$
         \quad for all $j$
  \State $s^h_j(t) \leftarrow \mathbf{1}[V^h_j(t) > \theta_h]$;
         \quad $V^h_j(t) \leftarrow 0$ if $s^h_j(t)=1$ \hfill (spike + reset)
  \State $V^o_k(t) \leftarrow \lambda\,V^o_k(t{-}1) + \sum_j W^2_{jk}\,s^h_j(t)$
         \quad for all $k$
  \State $s^o_k(t) \leftarrow \mathbf{1}[V^o_k(t) > \theta_o]$;
         \quad $V^o_k(t) \leftarrow 0$ if $s^o_k(t)=1$
\EndFor
 
\Statex \textbf{--- Readout \& reward ---}
\State $\hat{y} \leftarrow \arg\max_k \sum_t s^o_k(t)$
       \Comment{spike-count decision}
\If{reward mode $=$ \texttt{binary}}
  \State $r \leftarrow \mathbf{1}[\hat{y}=y] \cdot 2 - 1$
\ElsIf{reward mode $=$ \texttt{margin}}
  \State $r \leftarrow \bigl(\sum_t s^o_y(t) - \max_{k\neq y}\sum_t s^o_k(t)\bigr)\,/\,T$
\Else
  \State $r \leftarrow 1$ \Comment{no reward modulation}
\EndIf
 
\Statex \textbf{--- Stage 1: Output layer update ($\bm{W}_2$) ---}
\State $\tau_k(t) \leftarrow \mathbf{1}[k=y]$ for all $k, t$
       \Comment{target: correct class fires at all $T$ steps}
\State $\Delta W^2_{jk} \leftarrow \eta_2 \sum_t s^h_j(t)\,\bigl[\tau_k(t) - s^o_k(t)\bigr]$
       for all $j,k$
\State $\bm{W}_2 \leftarrow \gamma\,\bm{W}_2 + \Delta\bm{W}_2$;
       $\;\bm{W}_2 \leftarrow \mathrm{clip}(\bm{W}_2,\, -c,\, c)$
 
\Statex \textbf{--- Stage 2: Compute $K$-shifted agreement for each hidden neuron $j$ ---}
\State $s^{o,\star}(t) \leftarrow s^o_y(t)$ for $t=1,\ldots,T$
       \Comment{correct-class output spike train}
\For{$d = -K$ \textbf{to} $K$}
  \State Let $T_d = T - |d|$; extract overlapping segments
         $s^h_j[d]$, $s^{o,\star}[d]$ of length $T_d$
  \State Compute $p^h_j(d)$, $p^\star(d)$, $p^{\text{obs}}_j(d)$
         via Eqs.~\eqref{eq:p_h}--\eqref{eq:p_obs} on these segments
  \State $\kappa_j(d) \leftarrow
    \dfrac{p^{\text{obs}}_j(d) - p^e_j(d)}{1 - p^e_j(d)}$
    \quad where $p^e_j(d) = p^h_j(d)\,p^\star(d) + (1-p^h_j(d))(1-p^\star(d))$
\EndFor
\State $\tilde{\kappa}_j \leftarrow \dfrac{1}{2K+1}\sum_{d=-K}^{K}\kappa_j(d)$
       for all $j$
 
\Statex \textbf{--- Stage 2 (continued): Hidden layer update ($\bm{W}_1$) ---}
\State $\bar{x}_i \leftarrow \frac{1}{T}\sum_t s_i(t)$ for all $i$
\State $\Delta W^1_{ij} \leftarrow \eta_1\,\bar{x}_i\,\tilde{\kappa}_j\,r$
       for all $i,j$
\State $\bm{W}_1 \leftarrow \gamma\,\bm{W}_1 + \Delta\bm{W}_1$
\State $\mathbf{w}^1_j \leftarrow \mathbf{w}^1_j\,/\,\bigl(\|\mathbf{w}^1_j\|_2 + \varepsilon\bigr)$
       for all $j$
       \Comment{per-column $\ell_2$-normalisation}
\end{algorithmic}
\end{spacing}
\end{algorithm}
 
\paragraph{Comparison with STDP.}
Standard (unsupervised) STDP modifies a synapse based on the relative timing
of a single pre-synaptic and a single post-synaptic spike,
with an update magnitude that decays exponentially with their time difference.
The update is symmetric with respect to class labels: it is applied identically
whether the network's output is correct or not.
Reward-modulated STDP adds a third factor but retains the exponential-trace
mechanism, which requires tracking per-neuron trace variables at every
timestep and is sensitive to the choice of time constant $\tau$.
 
SADP differs in three ways.
First, the agreement measure $\kappa_j$ is \emph{computed over the entire
simulation window} as a summary statistic rather than propagated
step-by-step via decaying traces, eliminating the need to store and update
exponential state variables.
Second, the correct-class output spike train $s^{o,\star}$ provides an
\emph{explicit reference signal} tied to the label: the hidden neuron's
weight update is modulated by how well its activity correlates with the
network's own attempt to fire the correct class.
Third, the chance correction in $\kappa$ removes the confound of
coincidental agreement due to shared average firing rates, making the
agreement signal more informative per timestep.
The $K$-shift extension and reward modulation then add tolerance to temporal
jitter and trial-level feedback respectively, each as independent,
optionally-activated components.
 
\paragraph{Relation to biological plasticity and positioning.}
The SADP rule is a Hebbian rule in the following sense: no error gradient
is ever computed and no backward pass is required.
The output layer update is a standard pre$\times$post Hebbian rule with a
supervised target.
The hidden layer update uses pre-synaptic activity ($\bar{x}_i$) and a
post-synaptic agreement summary ($\tilde{\kappa}_j$), but the agreement is
measured against the \emph{correct-class output neuron's spike train}
$s^{o,\star}$, which originates from a different layer.
This cross-layer reference is not local in the strict synaptic sense,
the synapse $(i,j)$ does not have access to the output layer's activity
without some form of communication, but it requires only a single scalar
time series per timestep, rather than the weight-specific error signals
used by backpropagation.
The optional reward signal $r$ adds a global third factor analogous to
neuromodulatory gating in three-factor plasticity
rules~\cite{fremaux2016neuromodulated, kusmierz2017learning}, though the
primary supervisory signal in SADP is the forward-produced agreement
statistic rather than a reward signal.
The key algorithmic distinction from all existing approaches is that
supervision is delivered through a \emph{forward-computed, label-indexed
agreement measure} rather than through gradient propagation, reward
engineering, or teacher-injected currents.

\section{Experimental Details}
\label{sec:Experiments}

\subsection{Datasets}
\label{sec:datasets}

We evaluate on three standard image classification benchmarks and three
medical imaging datasets. The benchmark datasets: \textbf{MNIST}~\cite{lecun1998gradient},
\textbf{Fashion-MNIST (FMNIST)}~\cite{xiao2017fashion}, and
\textbf{CIFAR-10}~\cite{krizhevsky2009learning}, cover low- to
high-complexity visual inputs at fixed spatial resolutions of $28\times28$
(grayscale) and $32\times32$ (colour), with 10 classes each.
The medical datasets comprise \textbf{Colon Histopathology (COLON)} and
\textbf{Lung Histopathology (LUNG)}, both drawn from the LC25000
dataset~\cite{borkowski2019lc25000}, and \textbf{Brain MRI Tumor (TUMOUR)}~\cite{msoud_nickparvar_2021}.

Medical images were resized to $64\times64$ pixels using Lanczos resampling
and converted to three-channel (RGB) format prior to feature extraction.
All images were normalised to $[0,1]$.
The LC25000 data were partitioned into two sub-tasks: COLON (binary:
benign vs.\ adenocarcinoma; 10,000 images) and LUNG (three-class:
benign, adenocarcinoma, squamous cell carcinoma; 15,000 images).
The TUMOUR dataset~\cite{msoud_nickparvar_2021} was preprocessed via a
curated pipeline including deduplication, mislabel correction, resizing,
and augmentation (histogram equalisation, brightness adjustment, rotation,
flipping, and noise injection) prior to release; we used the distributed
version directly.
The dataset contains four categories: no tumour (normal), glioma tumour,
meningioma tumour, and pituitary tumour, for a total of 7{,}023 images.
Following the loader described in Section~\ref{sec:training_protocol}, all
images are merged and randomly partitioned 80/20 into 5{,}618 training and
1{,}405 test samples; the chance level for this four-class task is 25\%.
Table~\ref{tab:datasets} summarises key statistics.

\begin{table}[htbp]
\centering
\caption{Dataset statistics used in experiments.}
\label{tab:datasets}
\setlength{\tabcolsep}{5pt}
\begin{tabular}{@{}lcccccc@{}}
\toprule
\textbf{Dataset} & \textbf{Classes} & \textbf{Train} & \textbf{Test}
  & \textbf{Size} & \textbf{Channels} & \textbf{Type} \\
\midrule
MNIST      & 10 & 60,000 & 10,000 & $28\times28$ & 1 & Digit           \\
FMNIST     & 10 & 60,000 & 10,000 & $28\times28$ & 1 & Clothing        \\
CIFAR-10   & 10 & 50,000 & 10,000 & $32\times32$ & 3 & Object          \\
COLON      &  2 &  8,000 &  2,000 & $64\times64$ & 3 & Histopathology  \\
LUNG       &  3 & 12,000 &  3,000 & $64\times64$ & 3 & Histopathology  \\
TUMOUR     &  4 &  5{,}618 &  1{,}405 & $64\times64$ & 3 & Brain MRI \\
\bottomrule
\end{tabular}
\end{table}

\subsection{Input Encoding Strategies}
\label{sec:encodings}

All four encoding strategies map a normalised image (or its derived
feature vector) $\mathbf{x} \in [0,1]^{N_i}$ to a binary spike
train $s_i(t) \in \{0,1\}$ of length $T=50$ timesteps, using
independent Bernoulli sampling per neuron per timestep.
The encodings differ in how $\mathbf{x}$ is obtained; they are described
below in increasing order of discriminative complexity.

\subsubsection{Poisson-Only (Baseline)}
\label{sec:enc_poisson}

The image is flattened to a pixel intensity vector and used directly
as the Poisson rate vector: $s_i(t) \sim \mathrm{Bernoulli}(x_i)$.
No feature extraction is applied; every pixel drives one input neuron
independently.
This encoding tests whether the learning rule can extract discriminative
structure from raw pixel intensities, with no inductive bias.
The input dimension is $N_i = 784$ for MNIST and FMNIST ($28\times28$),
$N_i = 3{,}072$ for CIFAR-10 ($32\times32\times3$), and
$N_i = 12{,}288$ for the medical datasets ($64\times64\times3$).

\subsubsection{LBP + Poisson}
\label{sec:enc_lbp}

The Local Binary Pattern (LBP)~\cite{ojala2002multiresolution} descriptor
is computed on the grayscale version of the image using the standard
$P=8$ neighbour, $R=1$ radius configuration.
For each pixel at position $(y,x)$, the 8-bit LBP code is computed as
\begin{equation}
  \mathrm{LBP}(y,x) \;=\; \sum_{p=0}^{7} \mathbf{1}\!\left[g_p \geq g_c\right]\cdot 2^p,
  \label{eq:lbp}
\end{equation}
where $g_c$ is the intensity of the central pixel and $g_p$ denotes
the intensity of the $p$-th clockwise neighbour in the 8-ring.
The resulting code map (integer values in $[0,255]$) is divided into a
$4\times4$ spatial grid of non-overlapping blocks.
A 16-bin histogram of LBP codes is computed within each block and
$\ell_1$-normalised, yielding $4\times4\times16=256$ texture features.
For colour images (CIFAR-10 and all medical datasets), per-block
per-channel colour statistics (mean and clipped standard deviation for
each of the three RGB channels) are appended, adding
$4\times4\times3\times2=96$ dimensions for a total of $N_i=352$.
Grayscale inputs (MNIST, FMNIST) produce $N_i=256$.
All features are min-max normalised to $[0,1]$ across the training set
before Poisson encoding.

\subsubsection{LBP+CLBP + Poisson}
\label{sec:enc_clbp}

Complete Local Binary Pattern (CLBP)~\cite{ojala2002multiresolution}
decomposes the local neighbourhood into three complementary
binary codes over the same $P=8$, $R=1$ ring:
\begin{itemize}
  \item \textbf{CLBP\_S} (sign): the standard LBP code
        $\mathbf{1}[g_p \geq g_c]\cdot 2^p$, identical to Eq.~\eqref{eq:lbp}.
  \item \textbf{CLBP\_M} (magnitude): thresholds each absolute difference
        $|g_p - g_c|$ against the per-image mean absolute difference,
        yielding an 8-bit magnitude code.
  \item \textbf{CLBP\_C} (centre): a single bit per pixel indicating
        whether the central intensity exceeds the global image mean.
\end{itemize}
Block histograms of CLBP\_S and CLBP\_M (16 bins each over a
$4\times4$ grid) and CLBP\_C (2 bins) are computed and concatenated:
$(2\times4\times4\times16)+(4\times4\times2)=544$ features for
grayscale inputs ($N_i=544$ for MNIST, FMNIST),
with colour statistics appended for colour inputs to give $N_i=640$
(CIFAR-10 and medical datasets).
All features are min-max normalised identically to the LBP pipeline.
CLBP captures the magnitude of local intensity changes in addition to
their sign, providing richer texture statistics at approximately twice
the feature dimensionality of plain LBP, with no learnable parameters.

\subsubsection{CNN + Poisson}
\label{sec:enc_cnn}

A compact convolutional neural network (CNN) is pre-trained on the
labelled training set and used as a fixed feature extractor.
The architecture is:
\begin{center}
\textit{Conv}(32, $3\times3$, ReLU) $\to$ MaxPool($2\times2$)
$\to$ \textit{Conv}(64, $3\times3$, ReLU) $\to$ MaxPool($2\times2$)
$\to$ \textit{Conv}(128, $3\times3$, ReLU) $\to$ GlobalAvgPool
$\to$ \textit{Dense}(256, sigmoid)
\end{center}
All convolutions use same-padding.
The encoder is pre-trained end-to-end using a softmax classification head
appended to its output, optimised with Adam and sparse categorical
cross-entropy for 50 epochs (batch size 128, 10\% validation split).
After training the classification head is discarded; the 256-dimensional
sigmoid output of the Dense layer is used as the feature vector
($N_i=256$ for all datasets).
Features are min-max normalised per dimension using training-set
statistics before Poisson encoding.
The CNN encoder is re-trained independently for each algorithm (SADP and
STDP), dataset, and random seed, so neither algorithm benefits from a
shared feature representation.

Table~\ref{tab:encodings} summarises the four strategies.

\begin{table*}[htbp]
\centering
\small
\caption{Input encoding methods used in the experiments. $N_i$ denotes the SNN input dimension.
\textsuperscript{$\dagger$} For colour images; grayscale values are shown separately.}
\label{tab:encodings}

\setlength{\tabcolsep}{3pt}
\renewcommand{\arraystretch}{1.1}

\begin{tabular}{@{}l p{5.6cm} c c c@{}}
\toprule
\textbf{Encoding} &
\textbf{Description} &
\textbf{Train.} &
\textbf{Colour $N_i$} &
\textbf{Grey $N_i$} \\
\midrule

Poisson
&
Raw pixels converted to Bernoulli spike trains.
&
No
&
3072/12288
&
784
\\

LBP + Poisson
&
LBP histograms with colour moments.
&
No
&
352$^\dagger$
&
256
\\

LBP+CLBP + Poisson
&
CLBP histograms with colour moments.
&
No
&
640$^\dagger$
&
544
\\

CNN + Poisson
&
Pre-trained CNN features converted to Bernoulli spike trains.
&
Yes
&
\multicolumn{2}{c}{256}
\\

\bottomrule
\end{tabular}

\vspace{1mm}
\footnotesize
All feature vectors are converted to Bernoulli spike trains before being presented to the SNN. The CNN encoder produces a 256-dimensional feature vector for every dataset.

\end{table*}

\subsection{SNN Architecture and Neuron Parameters}
\label{sec:snn_arch}

All experiments use the single-hidden-layer architecture described in
Section~\ref{sec:network}: $N_i$ input neurons (encoding-dependent),
$N_h = 256$ hidden neurons, and $N_c$ output neurons equal to the number
of classes.
Weights are initialised as $W^1_{ij}, W^2_{jk} \sim \mathcal{N}(0, 0.1)$.

\paragraph{LIF neuron parameters.}
The membrane decay constant is $\lambda = 0.9$, corresponding to a
membrane time constant of approximately 9.5 timesteps.
To break symmetry, hidden neuron thresholds are drawn independently at
initialisation as $\theta_{h,j} \sim 0.5 + 0.05\,\mathcal{N}(0,1)$
and held fixed throughout training.
The output neuron threshold is $\theta_o = 0.5$ for all experiments.
The simulation window is $T = 50$ timesteps for all datasets and
both algorithms.

\paragraph{Prediction.}
The predicted class is determined by the spike-count readout
(Eq.~\ref{eq:readout}): $\hat{y} = \arg\max_k \sum_{t=1}^{T} s^o_k(t)$.

\subsection{Training Protocol}
\label{sec:training_protocol}

Both SADP and STDP are trained for 50 epochs with a mini-batch size of
128 and evaluated on the full test set after training.
The primary comparison sweep uses seed~42 for all configurations (controlling
weight initialisation, batch ordering, and spike sampling), enabling exact
reproduction.
All experiments are run on CPU; no GPU is required for the SNN training
stage, though the CNN pre-training step uses GPU-accelerated TensorFlow
when available.

To validate result stability across random initialisations, Supervised SADP
was additionally run with seeds $\{42, 43, 44\}$ on all three medical imaging
datasets (COLON, LUNG, TUMOUR) across LBP+Poisson, LBP+CLBP+Poisson, and
CNN+Poisson encodings.
Poisson-only was excluded from this multi-seed sweep due to its substantially
higher runtime on high-resolution medical images.
The benchmark datasets (MNIST, FMNIST, CIFAR-10) use fixed, globally
standardised train-test splits, so varying the seed affects only weight
initialisation and batch ordering but not the data partition; multi-seed
results for those datasets are therefore not reported separately.
Multi-seed results for medical datasets are presented in
Section~\ref{sec:multiseed}.

Weight regularisation is applied identically for both algorithms after
each mini-batch:
$\bm{W}_1$ columns are $\ell_2$-normalised (Eq.~\ref{eq:w1_norm})
following a multiplicative decay $\gamma = 0.9995$;
$\bm{W}_2$ receives the same decay followed by element-wise clipping
to $[-5, 5]$.

\subsection{SADP Hyperparameter Configuration}
\label{sec:sadp_config}

Table~\ref{tab:sadp_hparams} lists the full set of hyperparameters for
Supervised SADP.
The learning rates $\eta_1 = 2\times10^{-4}$ and $\eta_2 = 5\times10^{-4}$
are shared across all datasets and encodings without per-dataset tuning.

The two hyperparameters that are swept across the experimental grid are
$K_{\text{shift}}$ and the reward mode:

\paragraph{Shift parameter $K_{\text{shift}} \in \{5,\, 25\}$.}
$K=5$ allows the agreement measure to look $\pm5$ timesteps around each
comparison point, covering a temporal window of 11 steps;
$K=25$ extends this to $\pm25$ steps, covering the full half-window of
the simulation.
Setting $K=0$ (not evaluated here) recovers the synchronous rule of
Eq.~\eqref{eq:kappa} exactly.
Agreement at each offset $d$ is computed over the overlapping portion of
length $T_d = T - |d|$ and averaged uniformly over all $2K+1$ offsets
(Eq.~\ref{eq:kappa_shifted}).

\paragraph{Reward mode $\in \{\texttt{none},\,\texttt{binary},\,\texttt{margin}\}$.}
All three modes are evaluated for each $(K_{\text{shift}}, \text{encoding})$
combination, giving a $2\times3 = 6$-configuration inner sweep per
(dataset, encoding) cell.
With \texttt{none}, the reward factor $r=1$ for all samples
(Eq.~\ref{eq:w1_update}).
With \texttt{binary}, $r \in \{-1, +1\}$ based on whether the predicted
class matches the label (Eq.~\ref{eq:w1_reward}).
With \texttt{margin}, $r = (n_y - \max_{k\neq y}n_k)/T$ is the
normalised spike-count confidence margin (Eq.~\ref{eq:margin_reward}).

\begin{table}[!htbp]
\scriptsize
\centering
\caption{Supervised SADP hyperparameters. Fixed values are used across all datasets and encodings. Swept values define the experimental grid.}
\label{tab:sadp_hparams}

\setlength{\tabcolsep}{4pt}
\renewcommand{\arraystretch}{1.05}

\begin{tabular}{@{}lp{1.0cm}p{2.8cm}p{3.7cm}@{}}
\toprule
\textbf{Parameter} &
\textbf{Sym.} &
\textbf{Value} &
\textbf{Role} \\
\midrule

\multicolumn{4}{@{}l}{\textit{Fixed across all experiments}}\\

Simulation timesteps     & $T$          & 50                     & Spike train length\\
Hidden neurons           & $N_h$        & 256                    & Network width\\
Membrane decay           & $\lambda$    & 0.9                    & LIF leak constant\\
Hidden threshold         & $\theta_h$   & $0.5\pm0.05\mathcal N$ & Per-neuron, fixed after initialisation\\
Output threshold         & $\theta_o$   & 0.5                    & Output layer threshold\\
Hidden learning rate     & $\eta_1$     & $2\times10^{-4}$       & SADP update scale\\
Output learning rate     & $\eta_2$     & $5\times10^{-4}$       & Supervised Hebbian scale\\
Weight decay             & $\gamma$     & 0.9995                 & Batch-wise multiplicative decay\\
Output weight clip       & $c$          & 5.0                    & Element-wise bound on $\mathbf W_2$\\
$\ell_2$ normalisation   & $\varepsilon$& $10^{-6}$              & Column normalisation of $\mathbf W_1$\\
Training epochs          &              & 50                     & Per configuration\\
Batch size               &              & 128                    & Mini-batch size\\
Random seed              &              & 42; $\{42,43,44\}$     & Initialisation and data shuffling\\

\midrule
\multicolumn{4}{@{}l}{\textit{Swept hyperparameters}}\\

Temporal shift           & $K_{\rm shift}$ & $\{5,25\}$ & Agreement window\\
Reward mode              &                 & none, binary, margin & Third plasticity factor\\

\midrule
\multicolumn{4}{@{}l}{\textit{CNN encoder}}\\

Feature dimension        & $N_i$ & 256  & CNN output size\\
Pre-training epochs      &       & 50   & CNN training\\
Batch size               &       & 128  & Mini-batch size\\
Optimiser                &       & Adam & Default optimiser\\

\midrule
\multicolumn{4}{@{}l}{\textit{LBP/CLBP encoder}}\\

Neighbourhood            & $P,R$ & $8,1$    & Radius and neighbours\\
Spatial grid             &       & $4\times4$ & Non-overlapping blocks\\
Histogram bins           &       & 16       & Per-block histogram bins\\

\bottomrule
\end{tabular}
\end{table}

\subsection{STDP Baseline Configuration}
\label{sec:stdp_config}

The STDP baseline uses the same SNN architecture, LIF parameters, weight
regularisation, training protocol (50 epochs, batch size 128), and
identical encoding pipelines.
The output layer is updated with the same supervised Hebbian rule
(Eq.~\ref{eq:w2_update}) and the same reward mode as the SADP condition
it is compared against.
The hidden layer update follows the standard exponential-trace STDP rule:
the weight change for synapse $(i,j)$ is
\begin{equation}
  \Delta W^1_{ij} \;=\; \eta_1\,r\sum_{t=1}^{T}\!
  \Bigl[
    A_+\,x_i^+(t)\,s^h_j(t)
    \;-\;
    A_-\,x_j^-(t)\,s_i(t)
  \Bigr],
  \label{eq:stdp_update}
\end{equation}
where $x_i^+(t) = \sum_{t'<t} s_i(t')\,e^{-(t-t')/\tau_+}$ is the
pre-synaptic eligibility trace and
$x_j^-(t) = \sum_{t'<t} s^h_j(t')\,e^{-(t-t')/\tau_-}$ is the
post-synaptic trace, both decayed exponentially with their respective
time constants.
Reward modulation $r$ is applied identically to the SADP case.

The STDP-specific hyperparameters swept in the experimental grid are:

\paragraph{Time constant $\tau_+ = \tau_- \in \{2.0,\;10.0\}$ (timesteps).}
$\tau = 2.0$ gives a sharply decaying trace favouring tightly synchronised
pre–post pairs; $\tau = 10.0$ gives a broader eligibility window,
tolerating larger temporal offsets between pre-synaptic and post-synaptic
activity.
Setting $\tau_+ = \tau_-$ follows the symmetric convention commonly
used in supervised reward-modulated STDP~\cite{Mozafari2018, Deepr2024}.

\paragraph{Potentiation/depression amplitudes.}
$A_+ = A_- = 1.0$ for all experiments.
Asymmetric amplitude settings were not explored as the reward signal
already provides signed modulation.

Table~\ref{tab:stdp_hparams} summarises STDP-specific parameters.

\begin{table}[htbp]
\footnotesize
\centering
\caption{STDP baseline hyperparameters. Parameters shared with SADP
(network size, LIF dynamics, regularisation, training protocol,
encoding) are as listed in Table~\ref{tab:sadp_hparams}.}
\label{tab:stdp_hparams}
\setlength{\tabcolsep}{6pt}
\begin{tabular}{@{}llll@{}}
\toprule
\textbf{Parameter} & \textbf{Symbol} & \textbf{Value} & \textbf{Role} \\
\midrule
\multicolumn{4}{@{}l}{\textit{Fixed}} \\
LTP amplitude & $A_+$ & $1.0$ & Potentiation scale \\
LTD amplitude & $A_-$ & $1.0$ & Depression scale \\
Hidden learning rate & $\eta_1$ & $2\times10^{-4}$ & STDP update scale \\
\midrule
\multicolumn{4}{@{}l}{\textit{Swept}} \\
Trace time constant & $\tau_+ = \tau_-$ & $\{2.0,\;10.0\}$ & Eligibility window (timesteps) \\
Reward mode &   & \{\texttt{none}, \texttt{binary}, \texttt{margin}\} & Third factor \\
\bottomrule
\end{tabular}
\end{table}

\subsection{Experimental Grid and Evaluation}
\label{sec:exp_grid}

The full experimental grid spans:
6 datasets $\times$ 4 encodings $\times$ 2 hyperparameter values
($K_{\text{shift}}$ or $\tau$) $\times$ 3 reward modes $= 144$ configurations
per algorithm (288 total across SADP and STDP).
Feature extraction is computed once per (dataset, encoding, seed) triple
and cached; $K_{\text{shift}}$, $\tau$, and reward mode are varied at
SNN training time with no feature recomputation.

Performance is reported as test-set accuracy (\%) and macro-averaged F1
score (\%) computed on the held-out test split.
For each (dataset, encoding) pair, the best $K_{\text{shift}}$ (or $\tau$)
and reward mode are selected independently per algorithm by highest
test accuracy; the winning hyperparameter values are reported alongside
the result.
Full per-configuration results (all $K_{\text{shift}}$/reward-mode
combinations) are provided in Appendix~\ref{app:full_results}.

\section{Results}
\label{sec:results}
 
\begin{figure}[!htbp]
    \centering
    \includegraphics[width=\linewidth]{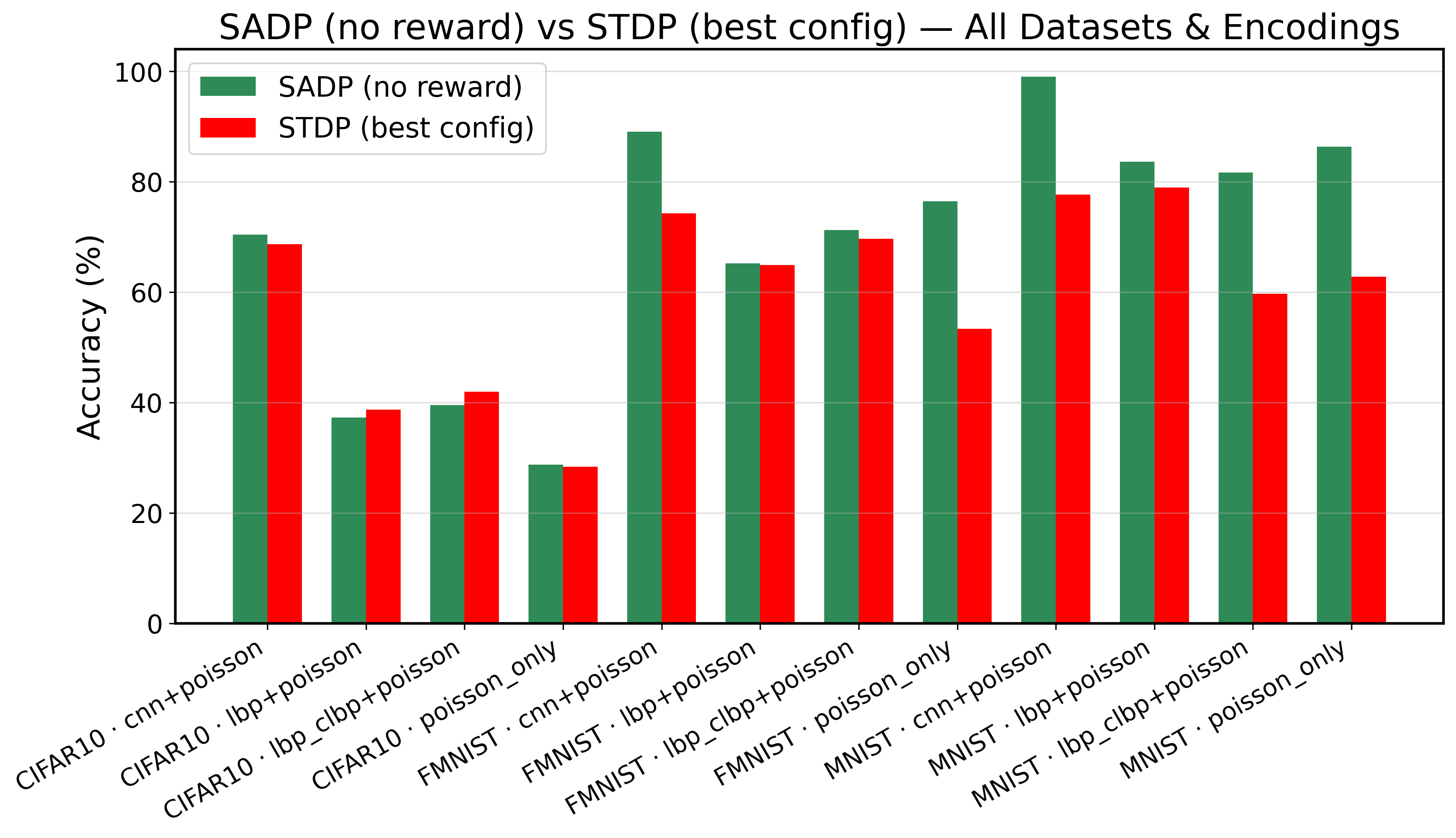}
    \caption{%
    Test accuracy of Supervised SADP in the reward-free \texttt{none} mode
    (green) versus the best-performing STDP configuration (red) for every
    benchmark dataset and encoding. Even without any reward signal, SADP
    matches or exceeds tuned STDP in most cells, with the largest margins
    under gradient-free Poisson and CNN$+$Poisson encodings on MNIST and
    Fashion-MNIST. The two CIFAR-10 LBP cells are the only benchmark cases
    where reward-modulated STDP is marginally ahead.}
    \label{fig:bar_overview}
\end{figure}

We compare Supervised SADP (\textbf{SADP}) against reward-modulated
Spike-Timing-Dependent Plasticity (\textbf{STDP}) across four input-encoding
strategies, three reward modulation modes (\emph{none}, \emph{binary},
\emph{margin}), two $K_{\text{shift}}$ values ($K\in\{5,25\}$) for SADP,
and two time-constant settings ($\tau\in\{2.0,10.0\}$\,ms) for STDP.
The complete sweep results are reported in ~\ref{app:full_results}.
In the following tables we report, for each (dataset, encoding) pair, the
best-performing configuration of each algorithm independently:
the selected $K$ or $\tau$ and reward mode are shown as separate columns so
that the winning hyperparameter is transparent.
\textbf{Bold} values indicate the higher-performing algorithm for that row.
Across the best configuration per
(dataset, encoding) pair, SADP leads in \textbf{10 of 12} benchmark
comparisons and \textbf{8 of 12} medical dataset comparisons in terms of classification accuracy.
A recurring theme throughout the results is that SADP with \texttt{none}
reward mode, relying solely on the Hebbian agreement with the correct-class
output spike train, often matches or outperforms STDP's best supervised
configuration, which requires a carefully chosen reward signal to achieve
class-specific learning.
This asymmetry directly reflects the structural difference between the two
learning rules: SADP's Hebbian term is inherently class-directed through
the forward-produced reference spike train, whereas STDP's Hebbian term
is class-agnostic and depends entirely on the reward gate for any
supervisory signal.
Figure~\ref{fig:bar_overview} visualises this comparison across all twelve
benchmark (dataset, encoding) cells, contrasting SADP in the reward-free
\texttt{none} mode against the best STDP configuration for each cell.

\subsection{Benchmark Image Classification Datasets}
\label{sec:results_benchmark}

Table~\ref{tab:benchmark} presents the best result per (dataset, encoding)
for both algorithms on MNIST, Fashion-MNIST (FMNIST), and CIFAR-10.
 
\paragraph{MNIST.}
SADP dominates across all four encodings.
The most algorithmically significant result is under Poisson-only encoding,
the fully gradient-free pipeline, where SADP achieves \textbf{86.46\%}
versus 62.80\% for the best STDP configuration, a gap of $+23.66$ pp,
without any CNN pre-training.
Under CNN$+$Poisson encoding (a high-end encoding baseline that includes
gradient-based CNN pre-training), SADP with $K{=}5$ and \emph{no reward
signal} (\texttt{none} mode) achieves \textbf{99.05\%} versus 77.66\% for
the best STDP configuration, a further $+21.39$ pp advantage.
In both cases the pattern is the same: SADP's agreement measure provides
sufficient supervisory signal without any external reward, while STDP
without reward modulation collapses to below 40\% accuracy
(see Appendix~\ref{app:full_results}).
This confirms that the agreement measure is inherently supervised: the
label enters through the forward-produced correct-class reference spike
train, not through an external reward signal.
LBP-based encodings yield consistent but smaller gains of $+4.85$ and
$+22.26$ pp respectively.
 
\paragraph{Fashion-MNIST.}
The pattern persists on the more challenging Fashion-MNIST dataset.
Under Poisson-only encoding SADP outperforms STDP by $+23.29$ pp
(\textbf{76.62\%} vs.\ 53.33\%), again with no CNN pre-training.
With CNN$+$Poisson the best SADP configuration ($K{=}25$, \emph{binary})
reaches \textbf{90.76\%} versus 74.25\% for STDP ($\tau{=}2.0$,
\emph{margin}), a $+16.51$ pp gap; even in the reward-free \texttt{none}
mode SADP achieves 89.08\%, still $+14.83$ pp above STDP's best.
LBP and LBP$+$CLBP encodings narrow the gap to $+0.44$ and $+1.69$ pp,
reflecting that hand-crafted texture features provide richer discriminative
structure that reduces the relative advantage of the agreement-based update.
For Fashion-MNIST the wider shift $K{=}25$ is selected for three of the
four encodings (Poisson-only, LBP$+$CLBP$+$Poisson, and CNN$+$Poisson),
with $K{=}5$ optimal only for LBP$+$Poisson, reflecting that the
higher-dimensional feature spaces benefit from a wider temporal window.
 
\paragraph{CIFAR-10.}
CIFAR-10 reveals a nuanced split.
Under Poisson-only encoding both algorithms perform poorly on this
complex colour dataset (SADP 29.90\% vs.\ STDP 28.38\%), confirming
that raw pixel Poisson encodings provide insufficient discriminative
structure for a single-layer SNN on CIFAR-10.
With LBP and LBP$+$CLBP encodings, STDP with $\tau{=}10.0$ and
\emph{margin} reward is marginally superior ($-1.43$ and $-2.32$ pp),
the only settings in the benchmark experiments where STDP outperforms SADP.
The STDP advantage with the larger time constant suggests that LBP features
produce longer-range temporal correlations that benefit from a wider
eligibility trace window.
With CNN$+$Poisson encoding SADP retains the advantage (\textbf{70.53\%},
$K{=}25$, \emph{margin} vs.\ 68.65\% for STDP, $+1.88$ pp), confirming
that the learning algorithm advantage holds across all encoding levels,
including the high-end CNN baseline.

\begin{table*}[htbp]
\scriptsize
\centering
\caption{%
  Benchmark dataset results (best per encoding for each algorithm).
  \emph{K}: best $K_{\text{shift}}$ for SADP.
  \emph{RM}: best reward mode (\texttt{n}~=~none, \texttt{b}~=~binary,
  \texttt{m}~=~margin).
  \emph{$\tau$}: best time constant for STDP.
  \emph{$t$/ep}: average time per training epoch~(s).
  \textbf{Bold} = better algorithm per row.
  $\Delta$ = SADP Acc.$-$STDP Acc.\ (pp).
  Full hyperparameter sweep in Appendix~\ref{app:sadp_bench} and
  \ref{app:stdp_bench}.%
}
\label{tab:benchmark}
\setlength{\tabcolsep}{4.2pt}
\renewcommand{\arraystretch}{1.05}
\begin{tabular}{@{}ll  cc cc c  cc cc c  r@{}}
\toprule
& &
\multicolumn{5}{c}{\textbf{SADP}} &
\multicolumn{5}{c}{\textbf{STDP}} & \\
\cmidrule(lr){3-7}\cmidrule(lr){8-12}
\textbf{Dataset} & \textbf{Encoding} &
  $K$ & RM & Acc.\% & F1\% & $t$/ep &
  $\tau$ & RM & Acc.\% & F1\% & $t$/ep &
  $\Delta$ \\
\midrule
\multirow{4}{*}{MNIST}
 & Poisson Only
   & 5 & b & \textbf{86.46} & \textbf{86.07} & \textbf{74.9}
   & 10.0 & b & 62.80 & 57.79 & 139.0 & $+23.66$ \\
 & LBP + Poisson
   & 5 & b & \textbf{83.76} & \textbf{83.73} & \textbf{35.6}
   & 2.0  & m & 78.91 & 78.39 & 52.3  & $+4.85$  \\
 & LBP+CLBP + Poisson
   & 5 & b & \textbf{81.92} & \textbf{81.63} & \textbf{57.2}
   & 2.0  & m & 59.66 & 51.24 & 101.0 & $+22.26$ \\
 & CNN + Poisson
   & 5 & n & \textbf{99.05} & \textbf{99.04} & \textbf{36.9}
   & 2.0  & m & 77.66 & 76.00 & 52.3  & $+21.39$ \\
\midrule
\multirow{4}{*}{FMNIST}
 & Poisson Only
   & 25 & b & \textbf{76.62} & \textbf{75.11} & \textbf{98.9}
   & 2.0  & m & 53.33 & 47.19 & 139.6 & $+23.29$ \\
 & LBP + Poisson
   & 5 & m & \textbf{65.33} & \textbf{61.60} & \textbf{35.6}
   & 2.0  & m & 64.89 & 61.12 & 52.1  & $+0.44$  \\
 & LBP+CLBP + Poisson
   & 25 & m & \textbf{71.35} & \textbf{68.62} & \textbf{80.8}
   & 2.0  & m & 69.66 & 66.01 & 98.9 & $+1.69$ \\
 & CNN + Poisson
   & 25 & b & \textbf{90.76} & \textbf{88.81} & 60.9
   & 2.0  & m & 74.25 & 74.11 & \textbf{52.0}  & $+16.51$ \\
\midrule
\multirow{4}{*}{CIFAR-10}
 & Poisson Only
   & 5 & b & \textbf{29.90} & \textbf{26.88} & \textbf{360.2}
   & 2.0  & m & 28.38 & 24.74 & 602.7 & $+1.52$  \\
 & LBP + Poisson
   & 5 & n & 37.30 & 33.62 & \textbf{35.5}
   & 10.0 & m & \textbf{38.73} & \textbf{35.30} & 56.8 & $-1.43$ \\
 & LBP+CLBP + Poisson
   & 5 & b & 39.64 & 36.30 & \textbf{52.6}
   & 10.0 & m & \textbf{41.96} & \textbf{38.63} & 95.2 & $-2.32$ \\
 & CNN + Poisson
   & 25 & m & \textbf{70.53} & \textbf{68.84} & 50.4
   & 2.0  & m & 68.65 & 67.20 & \textbf{44.0} & $+1.88$ \\
\bottomrule
\end{tabular}
\end{table*}

\subsection{Medical Imaging Datasets}
\label{sec:results_medical}

Table~\ref{tab:medical} presents results on the LC25000 colon histopathology
(COLON, binary), LC25000 lung histopathology (LUNG, three-class), and
Brain MRI Tumor (TUMOUR, four-class) datasets.
 
\paragraph{Colon Histopathology.}
SADP is superior on three of four encodings.
With Poisson-only encoding, SADP ($K{=}5$, \emph{binary}) achieves
\textbf{56.50\%} accuracy while the best STDP configuration ($\tau{=}10.0$,
\texttt{none}) degenerates to 50.00\% chance level for binary classification.
This collapse is consistent: STDP without reward modulation on raw Poisson
spikes from high-resolution histopathological images (input dimension 12,288)
cannot form discriminative synaptic weights from temporal correlations alone.
LBP and LBP$+$CLBP encodings bring both algorithms into a more comparable
regime ($\Delta{=}{+1.65}$ and $+0.75$ pp respectively), with the
lower-dimensional texture features providing richer statistical structure
for both learning rules.
CNN$+$Poisson encoding saturates the task for both algorithms:
SADP achieves 99.60\% and STDP achieves 99.80\%, a negligible $-0.20$ pp
difference that lies within run-to-run variability.
 
\paragraph{Lung Histopathology.}
SADP outperforms STDP across all four encodings on this three-class task.
The advantage is largest under Poisson-only ($+10.63$ pp,
\textbf{73.33\%} vs.\ 62.70\%) and LBP$+$CLBP$+$Poisson ($+6.40$ pp,
\textbf{80.07\%} vs.\ 73.67\%), confirming that the SADP agreement-based
update provides a larger advantage over STDP when the encoding provides
weaker discriminative structure.
With CNN$+$Poisson encoding SADP achieves \textbf{97.30\%} versus
96.93\% for STDP ($+0.37$ pp).
Notably, the optimal STDP for the CNN$+$Poisson case uses $\tau{=}2.0$
and \emph{binary} reward, whereas for LBP-based encodings the best STDP
configuration is $\tau{=}2.0$ with \emph{margin} reward, indicating
that the optimal reward mode for STDP changes with the encoding richness.
 
\paragraph{Brain MRI Tumor.}
TUMOUR is the most challenging dataset for both algorithms under
texture-based encodings.
Under Poisson-only, LBP, and LBP$+$CLBP encodings, both SADP and STDP
hover in the 51--56\% range (approximately double the 25\% chance level
for this four-class task), with STDP marginally leading in all three of
these non-CNN encodings ($-0.99$, $-0.81$, and $-2.24$ pp for Poisson-only,
LBP, and LBP$+$CLBP respectively).
This moderate performance reflects the high intra-class diversity and
complex texture patterns of the MRI images, which prevent a single-layer
SNN from reliably discriminating between the four tumour categories using
texture operators alone; both algorithms nonetheless learn partial
discriminative structure even without CNN features.
CNN$+$Poisson is the sole encoder that enables genuine learning.
Here SADP with $K{=}5$ and \texttt{none} reward achieves
\textbf{84.60\%} accuracy versus 83.61\% for STDP
($\tau{=}2.0$, \emph{margin}), a $+0.99$ pp advantage.
Critically, STDP with \texttt{none} reward and CNN$+$Poisson on TUMOUR
collapses to only 57.48\% (see Appendix~\ref{app:stdp_med}),
again demonstrating STDP's dependence on a carefully chosen reward signal
whereas SADP's agreement measure provides intrinsic supervision through
the forward-produced correct-class reference spike train.

\begin{table*}[htbp]
\scriptsize
\centering
\caption{%
  Medical imaging dataset results.
  Column notation identical to Table~\ref{tab:benchmark}.
  COLON is binary; LUNG is three-class; TUMOUR is four-class
  (glioma, meningioma, pituitary tumour, and no tumour; chance level 25\%).
  Full hyperparameter sweep in Appendix~\ref{app:sadp_med} and
  \ref{app:stdp_med}.%
}
\label{tab:medical}
\setlength{\tabcolsep}{4.2pt}
\renewcommand{\arraystretch}{1.05}
\begin{tabular}{@{}ll  cc cc c  cc cc c  r@{}}
\toprule
& &
\multicolumn{5}{c}{\textbf{SADP}} &
\multicolumn{5}{c}{\textbf{STDP}} & \\
\cmidrule(lr){3-7}\cmidrule(lr){8-12}
\textbf{Dataset} & \textbf{Encoding} &
  $K$ & RM & Acc.\% & F1\% & $t$/ep &
  $\tau$ & RM & Acc.\% & F1\% & $t$/ep &
  $\Delta$ \\
\midrule
\multirow{4}{*}{COLON}
 & Poisson Only
   & 5  & b & \textbf{56.50} & \textbf{50.96} & \textbf{172.7}
   & 10.0 & n & 50.00 & 33.33 & 483.4 & $+6.50$  \\
 & LBP + Poisson
   & 5  & b & \textbf{81.55} & \textbf{81.52} & 11.3
   & 2.0  & m & 79.90 & 79.85 & \textbf{8.8}  & $+1.65$  \\
 & LBP+CLBP + Poisson
   & 25 & n & \textbf{76.55} & \textbf{76.34} & 21.5
   & 2.0  & m & 75.80 & 75.63 & \textbf{14.8} & $+0.75$  \\
 & CNN + Poisson
   & 5  & m & 99.60 & 99.60 & \textbf{4.9}
   & 2.0  & m & \textbf{99.80} & \textbf{99.80} & 6.8  & $-0.20$  \\
\midrule
\multirow{4}{*}{LUNG}
 & Poisson Only
   & 25 & b & \textbf{73.33} & \textbf{70.43} & \textbf{254.8}
   & 2.0  & m & 62.70 & 52.39 & 640.3 & $+10.63$ \\
 & LBP + Poisson
   & 5  & n & \textbf{83.80} & \textbf{83.48} & \textbf{9.3}
   & 2.0  & m & 82.80 & 82.43 & 12.0  & $+1.00$  \\
 & LBP+CLBP + Poisson
   & 25 & m & \textbf{80.07} & \textbf{78.54} & \textbf{19.3}
   & 2.0  & m & 73.67 & 68.98 & 21.4  & $+6.40$  \\
 & CNN + Poisson
   & 5  & m & \textbf{97.30} & \textbf{97.29} & \textbf{7.7}
   & 2.0  & b & 96.93 & 96.93 & 9.2   & $+0.37$  \\
\midrule
\multirow{4}{*}{TUMOUR}
 & Poisson Only
   & 25 & b & 51.58 & 40.97 & \textbf{377.3}
   & 2.0  & b & \textbf{52.57} & \textbf{41.89} & 1078.0 & $-0.99$  \\
 & LBP + Poisson
   & 25 & n & 51.65 & 41.59 & 22.6
   & 10.0 & b & \textbf{52.46} & \textbf{43.09} & \textbf{18.9} & $-0.81$ \\
 & LBP+CLBP + Poisson
   & 5  & b & 53.42 & 45.93 & \textbf{20.5}
   & 10.0 & b & \textbf{55.66} & \textbf{50.78} & 31.4  & $-2.24$  \\
 & CNN + Poisson
   & 5  & n & \textbf{84.60} & \textbf{84.59} & \textbf{11.1}
   & 2.0  & m & 83.61 & 83.74 & 13.3  & $+0.99$  \\
\bottomrule
\end{tabular}
\end{table*}

\subsection{Training Efficiency}
\label{sec:results_speed}
 
We quantify training efficiency with a single, fully reproducible metric.
For each (dataset, encoding) pair we take the \emph{best-accuracy}
configuration of each algorithm, exactly the configurations already
reported in Tables~\ref{tab:benchmark} and~\ref{tab:medical} and define
the speedup as the ratio of wall-clock time per epoch,
\begin{equation}
  s_{d,e} \;=\; \frac{t^{\text{STDP}}_{d,e}}{t^{\text{SADP}}_{d,e}},
  \label{eq:speedup}
\end{equation}
where $t^{\text{STDP}}_{d,e}$ and $t^{\text{SADP}}_{d,e}$ are the $t$/ep
values of those same best-accuracy configurations (the $K$/$\tau$ and
reward mode are listed alongside each ratio in
Table~\ref{tab:speed_full}). This yields one speedup per cell: 24 in
total, six datasets $\times$ four encodings, and values $s_{d,e}>1$ mean
SADP trains faster. As a worked example, on MNIST with Poisson-only
encoding SADP's best configuration ($K{=}5$, \texttt{binary}) runs at
$74.9$\,s/epoch and STDP's best ($\tau{=}10.0$, \texttt{binary}) at
$139.0$\,s/epoch, so $s = 139.0/74.9 = 1.86\times$.
Per-cell speedups appear in Table~\ref{tab:speed_full} and the
per-encoding summary in Table~\ref{tab:speed}.
 
Averaged over all 24 (dataset, encoding) cells, SADP trains
$1.47\times$ faster than STDP and is faster in \textbf{19 of 24} cells.
The largest and most consistent gains occur with Poisson-only encoding
($2.14\times$ average, faster in all six cells), where STDP's trace-decay
computation is most expensive relative to the signal extracted: it must
maintain exponentially-decaying eligibility traces over all $T{=}50$
timesteps for every pre-/post-synaptic spike pair in a high-dimensional
Poisson spike train, whereas SADP computes a single window-averaged
agreement statistic per hidden neuron. CNN$+$Poisson shows the smallest
average gain ($1.15\times$) because the dominant cost, the CNN feature
extraction, is shared by both algorithms, so only the small learning-rule
overhead differs. The largest single speedup is $2.86\times$ on TUMOUR with
Poisson-only encoding, where STDP requires $1078.0$\,s/epoch against
$377.3$\,s for SADP.
 
SADP is slower in only 5 of 24 cells (range $0.69$--$0.87\times$). Two are
benchmark CNN$+$Poisson cells (FMNIST $0.85\times$, CIFAR-10 $0.87\times$)
where SADP's best-accuracy configuration uses the wider shift $K{=}25$,
which adds forward-pass overhead relative to STDP's inexpensive
$\tau{=}2.0$ update. The other three are small-image medical texture cells
(COLON LBP $0.78\times$, COLON LBP$+$CLBP $0.69\times$, TUMOUR LBP
$0.84\times$), where the low spatial resolution makes STDP's per-update
trace maintenance cheap while SADP's shift computation does not
proportionally shrink. In every high-dimensional or Poisson-encoded
setting, precisely the regime where per-epoch cost dominates total
training time, SADP is the faster rule.
 
\begin{table}[htbp]
\centering
\caption{%
  Per-encoding training speedup, computed by Eq.~\eqref{eq:speedup}:
  for each (dataset, encoding) the ratio
  $t^{\text{STDP}}/t^{\text{SADP}}$ of the best-accuracy configurations
  reported in Tables~\ref{tab:benchmark} and~\ref{tab:medical}, then
  averaged over the six datasets. \textbf{Min}/\textbf{Max} are over the
  same six per-encoding cells; \textbf{SADP faster} counts cells with
  $s>1$. Values $>1$ indicate SADP is faster. Per-cell values in
  Table~\ref{tab:speed_full}.%
}
\label{tab:speed}
\setlength{\tabcolsep}{8pt}
\begin{tabular}{@{}lcccc@{}}
\toprule
\textbf{Encoding} & \textbf{Avg Speedup} & \textbf{Min} & \textbf{Max} & \textbf{SADP faster} \\
\midrule
Poisson Only         & $2.14\times$ & $1.41\times$ & $2.86\times$ & 6/6 \\
LBP + Poisson        & $1.24\times$ & $0.78\times$ & $1.60\times$ & 4/6 \\
LBP+CLBP + Poisson   & $1.35\times$ & $0.69\times$ & $1.81\times$ & 5/6 \\
CNN + Poisson        & $1.15\times$ & $0.85\times$ & $1.42\times$ & 4/6 \\
\midrule
\textbf{Overall}     & $\mathbf{1.47\times}$ & $0.69\times$ & $2.86\times$ & \textbf{19/24} \\
\bottomrule
\end{tabular}
\end{table}

\subsection{Hyperparameter Analysis}
\label{sec:results_hyperparam}

\paragraph{SADP: shift parameter $K$.}
The smaller shift $K{=}5$ is optimal in the majority of configurations.
Examining the 24 best-per-(dataset, encoding) selections across all
experiments, $K{=}5$ is selected in 15 cases and $K{=}25$ in 9.
$K{=}25$ is preferred when input representations are high-dimensional
(Poisson-only on LUNG, TUMOUR) or when combined texture features
(LBP$+$CLBP) produce a richer spike-time structure that benefits from
a wider temporal agreement window.
The performance gap between $K{=}5$ and $K{=}25$ is small in most
cases (typically $<\!1$ pp; see Appendix~\ref{app:sadp_bench}),
confirming that SADP is relatively robust to this hyperparameter
once a coarse regime is selected.
 
\paragraph{STDP: time constant $\tau$.}
STDP is considerably more sensitive to $\tau$.
The shorter constant $\tau{=}2.0$\,ms is selected in 18 of 24 best
configurations, with $\tau{=}10.0$\,ms appearing in the remaining six:
MNIST and COLON under Poisson-only encoding, and CIFAR-10 and TUMOUR
under LBP-based encodings.
More critically, the \emph{combination} of $\tau$ and reward mode is
decisive: several $(\tau, \text{RM})$ pairs collapse to near-random
accuracy even when the complementary pair succeeds.
For example, MNIST CNN$+$Poisson with $(\tau{=}2.0, \texttt{none})$
achieves only 38.03\%, and with $(\tau{=}10.0, \texttt{none})$ only
27.48\%, whereas $(\tau{=}2.0, \texttt{margin})$ reaches 77.66\%.
This sensitivity demands careful joint tuning of $\tau$ and reward
mode and represents a practical deployment challenge absent from SADP.

\subsection{Multi-Seed Stability on Medical Imaging Datasets}
\label{sec:multiseed}

The benchmark datasets (MNIST, FMNIST, CIFAR-10) use globally standardised
fixed train-test splits, so seed variation affects only weight initialisation
and batch ordering, not the data partition; the single-seed results for those
datasets are therefore representative in the same sense as comparable
single-run benchmarks in the SNN literature.
Thus, to assess result stability across random initialisations, Supervised SADP was
re-run with seeds $\{42, 43, 44\}$ on all three medical imaging datasets using
LBP+Poisson, LBP+CLBP+Poisson, and CNN+Poisson encodings. Full multi-seed validation of the SADP vs.\ STDP comparison across all six
datasets and all encodings remains computationally expensive, particularly
for Poisson-only encodings on high-resolution images.
Table~\ref{tab:multiseed} reports the mean and standard deviation of test
accuracy and F1 score across the three seeds for the best-performing
configuration per (dataset, encoding) combination.

CNN+Poisson encoding shows the highest stability across all three datasets:
standard deviations of at most $0.05$ pp for COLON, $0.71$ pp for LUNG, and
$0.99$ pp for TUMOUR.
This consistency reflects the dominant contribution of the fixed CNN feature
extractor: once discriminative features are extracted, the SADP learning rule
converges reliably regardless of weight initialisation.
LBP-based encodings exhibit somewhat higher variability ($0.62$--$2.28$ pp),
which is expected because the SNN must learn feature-to-class mappings
entirely through the local Hebbian update without a pre-trained feature
representation.
The moderate accuracy on TUMOUR with texture-based encodings
($\approx 49$--54\%, approximately twice the 25\% chance level for this
four-class task) is accompanied by larger standard deviations
($1.63$--$2.28$ pp), reflecting the genuine difficulty of discriminating
between tumour subtypes rather than algorithmic instability: both the mean
and the spread confirm that neither $K_{\text{shift}}$ value nor reward mode
produces a reliable learning signal on this dataset without CNN features.
Across all configurations where the mean accuracy is substantially above
chance (COLON and LUNG with any encoding; TUMOUR with CNN+Poisson), the
standard deviation is at most $1.77$ pp, indicating that the reported
single-seed results in Tables~\ref{tab:medical} are
representative of the algorithm's typical behaviour.

\begin{table}[htbp]
\scriptsize
\centering
\caption{%
  Multi-seed stability of Supervised SADP on medical imaging datasets
  (seeds $\{42,43,44\}$, $n{=}3$).
  Results show the mean$\,\pm\,$standard deviation of test accuracy and
  macro-F1 (\%) across seeds for the best-performing configuration per
  (dataset, encoding).
  $K$ and RM (reward mode; n=none, b=binary, m=margin) denote the
  configuration achieving the highest mean accuracy.
  Poisson-only encoding excluded due to runtime constraints.%
}
\label{tab:multiseed}
\setlength{\tabcolsep}{5pt}
\renewcommand{\arraystretch}{1.05}
\begin{tabular}{@{}llcc cc@{}}
\toprule
\textbf{Dataset} & \textbf{Encoding}
  & $K$ & RM
  & \textbf{Acc.\% (mean$\,\pm\,$std)}
  & \textbf{F1\% (mean$\,\pm\,$std)} \\
\midrule
\multirow{3}{*}{COLON}
  & CNN + Poisson       & 5  & m & $99.80 \pm 0.05$ & $99.80 \pm 0.05$ \\
  & LBP + Poisson       & 5  & b & $81.52 \pm 0.70$ & $81.48 \pm 0.73$ \\
  & LBP+CLBP + Poisson  & 5  & m & $77.92 \pm 1.40$ & $77.79 \pm 1.46$ \\
\midrule
\multirow{3}{*}{LUNG}
  & CNN + Poisson       & 5  & n & $97.59 \pm 0.71$ & $97.59 \pm 0.71$ \\
  & LBP + Poisson       & 5  & n & $83.44 \pm 1.53$ & $83.01 \pm 1.77$ \\
  & LBP+CLBP + Poisson  & 5  & n & $79.14 \pm 0.62$ & $77.26 \pm 0.86$ \\
\midrule
\multirow{3}{*}{TUMOUR}
  & CNN + Poisson       & 5  & b & $83.83 \pm 0.99$ & $84.09 \pm 1.04$ \\
  & LBP + Poisson       & 5  & n & $49.19 \pm 2.28$ & $40.45 \pm 3.34$ \\
  & LBP+CLBP + Poisson  & 5  & b & $53.82 \pm 1.63$ & $46.71 \pm 1.88$ \\
\bottomrule
\end{tabular}
\end{table}

\section{Discussion}
\label{sec:discussion}

\begin{figure*}[!htbp]
    \centering
    \includegraphics[width=0.92\linewidth]{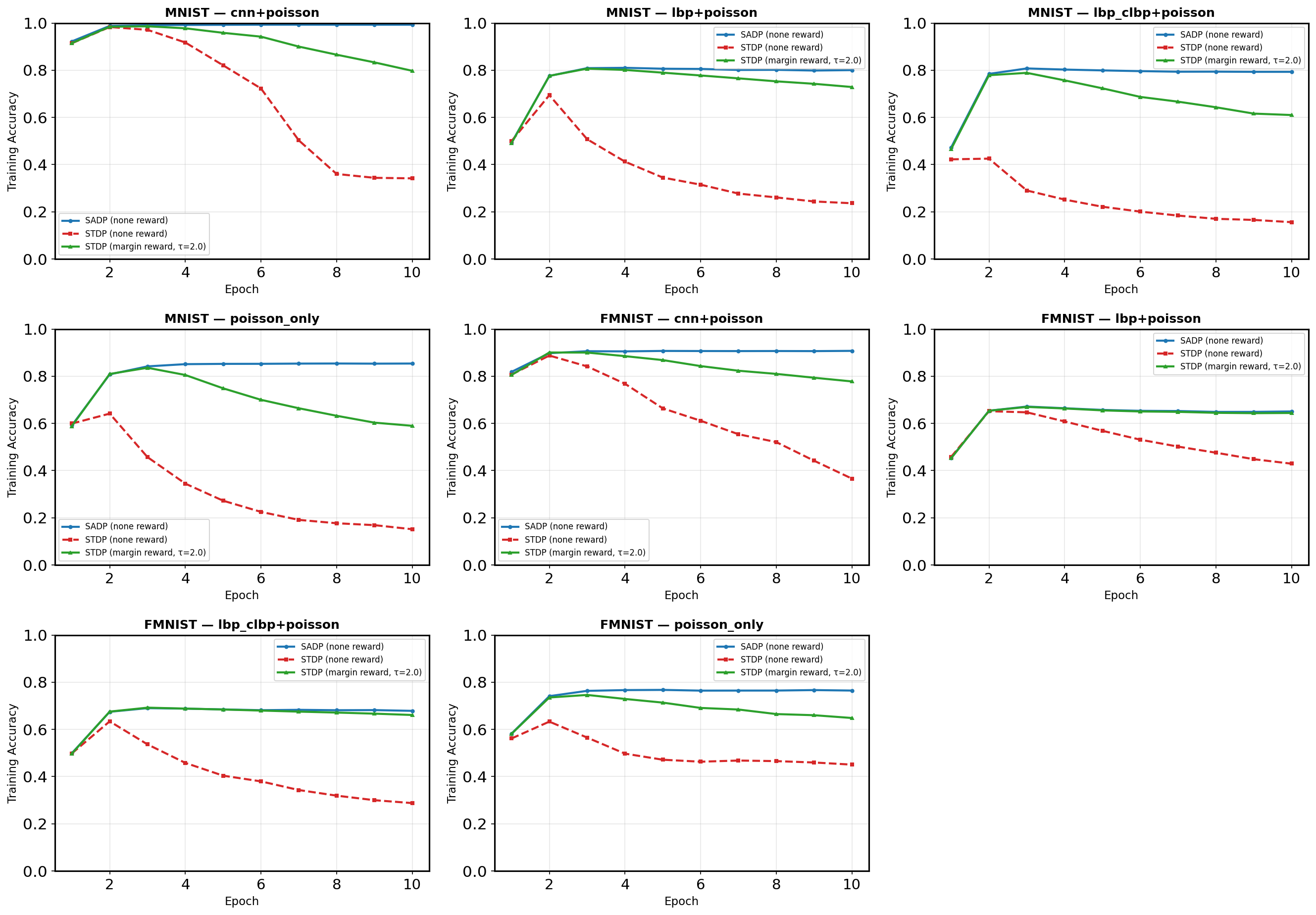}
    \caption{%
    Epoch-wise training accuracy on MNIST (top row) and Fashion-MNIST
    (bottom row) across the four encodings, comparing Supervised SADP
    (\texttt{none} reward), STDP (\texttt{none} reward), and the
    best reward-modulated STDP (margin reward, $\tau=2.0$). Reward-free SADP
    converges rapidly and holds its accuracy, whereas reward-free STDP
    collapses after an early peak; margin-reward STDP recovers much of the
    gap but remains below SADP and less stable, illustrating STDP's
    structural dependence on an external reward signal.}
    \label{fig:learning_curves}
\end{figure*}

\paragraph{The Hebbian agreement signal encodes supervision without any external reward.}
The central finding of this work is that Supervised SADP with \texttt{none}
reward mode relying solely on the Hebbian agreement between each hidden
neuron and the forward-produced correct-class output spike train learns
effective classifiers without any reward circuit.
The most direct demonstration is under fully gradient-free Poisson-only
encoding, where SADP achieves \textbf{86.46\%} on MNIST and
\textbf{76.62\%} on Fashion-MNIST, outperforming the best STDP configuration
(which requires a reward signal) by 23.66 and 23.29 percentage points
respectively, with no CNN pre-training; in the reward-free \texttt{none}
mode SADP reaches essentially the same accuracy (86.37\% and 76.59\%),
confirming that the agreement signal alone carries the supervision.
Even with CNN$+$Poisson encoding, SADP with \texttt{none} reward achieves
\textbf{99.05\%} on MNIST versus 77.66\% for STDP's best supervised
configuration, a $+21.39$ pp advantage with an identical architecture.
The reason is mechanistic: in SADP, the Hebbian correlation term is computed
between each hidden neuron and the \emph{correct-class output neuron's} spike
train, which is itself shaped by the supervised Hebbian update on $\bm{W}_2$.
Class label information therefore enters the hidden layer update directly,
through the reference spike pattern $s^{o,\star}(t) = s^o_y(t)$, without any
external reward computation.
In STDP, the pre--post timing correlation has no reference to the correct
class; without reward modulation it is functionally unsupervised, and the
results confirm this, STDP with \texttt{none} mode degenerates to between
roughly 10\% (near the 10-class chance level) and 49\% accuracy across
datasets and time constants in the MNIST, FMNIST and CIFAR-10 datasets.
This is because,
SADP's Hebbian term is class-directed by construction; STDP's is not.
This divergence is evident in the epoch-wise training dynamics of
Figure~\ref{fig:learning_curves}: reward-free SADP rises quickly and remains
stable, reward-free STDP peaks early and then steadily deteriorates as its
class-agnostic correlations accumulate, and only reward-modulated STDP
tracks SADP, and even then less stably on several encodings.

\paragraph{Encoding quality determines the performance ceiling for both algorithms.}
Across all six datasets the CNN+Poisson encoding yields the highest absolute
accuracy for both SADP and STDP in every case where the task is not already
saturated.
This is expected: the CNN pre-training provides a 256-dimensional feature
space that concentrates class-discriminative variation, making the
single-layer SNN's learning problem substantially simpler. But this is backpropagation-assisted performance.

On the raw Poisson-only encoding, where each pixel drives one input neuron
independently, both algorithms struggle on CIFAR-10 (roughly 28–30\%)
and on the medical datasets without CNN (roughly 50–73\%), confirming that
a single spiking hidden layer cannot learn sufficient intermediate
representations from pixel-level Poisson spikes on complex images.
This finding should be interpreted carefully: the strong SADP results on
CNN+Poisson do not demonstrate that SADP alone solves complex classification,
but rather that a well-designed hybrid pipeline combining discriminative
feature extraction with gradient-free Hebbian learning can do so.
The CNN pre-training stage uses gradient descent and is not gradient-free;
the gradient-free property of SADP applies specifically to the SNN training
stage that follows feature extraction.

\paragraph{STDP's sensitivity to hyperparameter configuration is a practical concern.}
The STDP results reveal a strong dependency on the joint choice of time
constant $\tau$ and reward mode.
On the TUMOUR dataset with CNN+Poisson encoding, STDP's accuracy ranges from
37.09\% ($\tau{=}10.0$, \texttt{none}) to 83.61\% ($\tau{=}2.0$,
\texttt{margin}), a 46.52 percentage point spread across the six
configurations of the same algorithm on identical data.
Similarly, on MNIST CNN+Poisson, STDP with \texttt{none} reward achieves
only 27.5–38.0\% depending on $\tau$, while the best STDP configuration
reaches 77.7\%.
SADP, by contrast, spans at most 0.28 percentage points across reward modes
on any single dataset (on the best encoding), and the two $K$ values
produce comparable results in most configurations.
This asymmetry in sensitivity is practically significant: deploying STDP
requires careful joint tuning of $\tau$ and reward mode, while SADP's
agreement-based plasticity provides more predictable behaviour across
configurations.
This observation does not imply that SADP has no hyperparameters requiring
selection, $K_{\text{shift}}$ must still be chosen, but the performance
consequences of a suboptimal $K$ are smaller than those of a suboptimal
$(\tau, \text{reward})$ pair for STDP.

\paragraph{The CIFAR-10 LBP exception reveals complementary properties of the two rules.}
On CIFAR-10 with LBP and LBP+CLBP encodings, STDP outperforms SADP by up
to 2.5 percentage points under binary and margin reward modes.
Examining the reward-mode breakdown clarifies the structure of this
exception: with \texttt{none} reward, STDP collapses to approximately 11\%
on CIFAR-10 LBP+CLBP (essentially the 10-class chance level), while SADP
reaches 39.5\%, a $+28.6$ pp advantage.
The STDP advantage appears only when reward modulation is enabled, and is
largest with the wider $\tau=10.0$ eligibility window, i.e., when a wide
eligibility trace window is combined with a global correction signal.
The LBP+CLBP features on CIFAR-10 produce rich multi-scale texture
histograms with longer-range local correlations; a wider STDP trace
($\tau=10$) can exploit these correlations in a way that SADP's fixed
$K$-shift window does not.

\paragraph{Limitations.}
Several limitations of this study should be acknowledged.
First, the network architecture is a single-hidden-layer SNN with
$N_h{=}256$ neurons; whether Supervised SADP extends gracefully to deeper
networks, recurrent connections, or convolutional spiking layers is an open
question that this work does not address.
Second, the gradient-free claim applies to the SNN training stage only;
the CNN+Poisson pipeline requires a gradient-based pre-training step,
so the overall system is not fully gradient-free for those experiments.
Poisson-only and LBP-based pipelines involve no gradient computation at any
stage, but their absolute accuracy lags CNN+Poisson significantly on complex
datasets.
Third, no comparison with surrogate-gradient or e-prop trained SNNs is
included; such methods would likely yield higher absolute accuracy on
all tasks at the cost of requiring backpropagation. In this work we deliberately avoid using backpropagation based training as much as we could. The CNN+Poisson case does use backpropagation but for independent feature generation and not as a part of the SNN classifier directly.

\paragraph{Implications and contribution to the community.}
This work makes three contributions beyond the specific algorithm evaluated.
\textbf{Conceptually}, we show that class-label supervision can be embedded
directly in the Hebbian plasticity term, through agreement with a
forward-produced class-specific spike pattern, with no reward signal,
teacher current, or gradient. Supervision thus becomes a matter of
\emph{which reference neuron} the Hebbian correlation is computed against,
rather than something bolted onto an unsupervised rule; concretely, SADP
learns effectively in the \texttt{none} reward mode, whereas STDP degrades
sharply, often to near-chance, without an engineered reward.
\textbf{Practically}, SADP offers a distinct trade-off: unlike
surrogate-gradient methods it needs no backward pass (incompatible with
hardware lacking backward connectivity~\cite{davies2018loihi}), and unlike
reward-modulated STDP it needs no reward engineering, while using an
$\mathcal{O}(T)$ update and training faster than STDP in most configurations.
A device-calibrated run using a measured iontronic memtransistor conductance
kernel (Appendix~\ref{app:device}) stays within $1.5$\,pp of the
standard algorithm, evidencing hardware compatibility. \textbf{Empirically},
the full sweep over six datasets, four encodings, and two algorithms provides
a transparent, reproducible baseline, and framing plasticity via
Cohen's~$\kappa$ invites future rules built on other association measures
(mutual information, rank correlation).
\section{Conclusion}
\label{sec:conclusion}

We introduced Supervised SADP, a gradient-free, backpropagation-free
Hebbian learning algorithm for spiking neural networks that extends the
unsupervised SADP framework~\cite{bej2025sadp} to supervised classification.
The algorithm consists of two sequential gradient-free updates: a supervised
Hebbian rule that encodes class labels into the output spike pattern,
followed by a chance-corrected spike agreement update (Cohen's $\kappa$)
that modulates hidden-layer weights according to each neuron's temporal
agreement with the correct-class output spike train produced by the forward
pass.
A $K$-shift extension aggregates agreement over a temporal window of
configurable width, and an optional reward-modulation term provides a global
third plasticity factor, both as independently composable additions to the
base rule.

Evaluated across six datasets and four input encodings, Supervised SADP
outperforms reward-modulated STDP in a significant majority of encoding–reward-mode
comparisons.
Under fully gradient-free Poisson-only encoding, SADP outperforms the best
STDP configuration by 23.66 pp on MNIST and 23.29 pp on Fashion-MNIST;
the advantage holds in the large majority of cells, including under
CNN-extracted features as a high-end encoding baseline.
The key distinguishing property is that SADP's Hebbian term is class-directed
by construction: the agreement with the correct-class output spike train
carries label information directly into the weight update, without any
reward gate.
This makes SADP effective with \texttt{none} reward mode on most tasks,
whereas STDP without a reward signal degrades sharply, often to near-chance
performance, a structural difference, not merely a quantitative one.
STDP's performance is strongly contingent on the joint choice of time
constant and reward mode, collapsing in many configurations where SADP
remains stable.
SADP is also faster, training at $1.47\times$ the speed of STDP on
average and up to $2.86\times$ faster under high-dimensional Poisson
encodings.

The primary limitation of the current study is the single-hidden-layer
architecture; future work should assess Supervised SADP in deeper SNN
architectures and investigate whether the agreement-based update can be
extended to recurrent or convolutional spiking networks without requiring
gradient computation.
Multi-seed validation has been demonstrated for medical datasets
(Section~\ref{sec:multiseed}); extending this to the full SADP--STDP
comparison across all datasets and encodings remains computationally
expensive and is deferred to future work.
Fully gradient-free pipelines, using Poisson or LBP encodings rather
than CNN pre-training remain a meaningful target for tasks where
hardware constraints preclude any gradient computation at any stage.

Supervised SADP occupies a specific and currently underserved design
point in the landscape of SNN learning: gradient-free, reward-engineering-free,
and supervised simultaneously, achieved not by adding supervision on top
of an unsupervised Hebbian rule, but by making the Hebbian term itself
class-directed.
We hope the algorithm, the evaluation framework, and the publicly
released hyperparameter sweep results provide a useful reference point
for the neuromorphic learning community as it develops practical training
methods for spike-based computation.

\section*{Code Availability}

The implementation of Supervised Spike Agreement-Dependent Plasticity (Supervised SADP), together with all code required to reproduce the results presented in this work, is publicly available at
\url{https://github.com/gouri88/Supervised-SADP-}.



\bibliographystyle{elsarticle-num}
\bibliography{cas-refs}

\appendix

\section{Complete Hyperparameter Sweep Results}
\label{app:full_results}
 
The following four tables
(Tables~\ref{tab:app_sadp_bench}, \ref{tab:app_stdp_bench},
\ref{tab:app_sadp_med}, and~\ref{tab:app_stdp_med})
report results for every combination of
$K_{\text{shift}}\in\{5,25\}$ (SADP) or $\tau\in\{2.0,10.0\}$\,ms (STDP)
and reward mode $\in\{\texttt{none},\texttt{binary},\texttt{margin}\}$
for all encoding strategies and datasets.
All accuracy and F1 values are in percent (\%).
$t$/ep denotes the average wall-clock time per training epoch in seconds.
Results are sorted by Dataset $\to$ Encoding $\to$ Hyperparameter $\to$
Reward Mode.

\subsection{SADP: Benchmark Datasets (Full Sweep)}
\label{app:sadp_bench}

\begin{longtable}{@{}llll rrrr r@{}}
\caption{%
  SADP full hyperparameter sweep: MNIST, FMNIST, CIFAR-10.
  $K$: shift parameter. RM: reward mode (n=none, b=binary, m=margin).%
}
\label{tab:app_sadp_bench} \\
\toprule
\textbf{Dataset} & \textbf{Encoding} & $K$ & \textbf{RM}
  & \textbf{Acc.\%} & \textbf{F1\%} & \textbf{Prec.\%} & \textbf{Rec.\%}
  & \textbf{$t$/ep} \\
\midrule
\endfirsthead
\multicolumn{9}{l}{\small\itshape (continued from previous page)} \\
\toprule
\textbf{Dataset} & \textbf{Encoding} & $K$ & \textbf{RM}
  & \textbf{Acc.\%} & \textbf{F1\%} & \textbf{Prec.\%} & \textbf{Rec.\%}
  & \textbf{$t$/ep} \\
\midrule
\endhead
\midrule
\multicolumn{9}{r}{\small\itshape (continued on next page)} \\
\endfoot
\bottomrule
\endlastfoot

\multirow{12}{*}{MNIST}
 & \multirow{6}{*}{\makecell[l]{CNN +\\Poisson}}
   & 25 & n & 98.99 & 98.98 & 98.99 & 98.97 & 61.8 \\
 & & 25 & b & 98.99 & 98.98 & 98.99 & 98.97 & 61.8 \\
 & & 25 & m & 99.00 & 98.99 & 99.00 & 98.98 & 61.8 \\
 & & 5  & n & \textbf{99.05} & \textbf{99.04} & 99.04 & 99.03 & 36.9 \\
 & & 5  & b & 99.00 & 99.00 & 99.00 & 99.00 & 37.1 \\
 & & 5  & m & 99.04 & 99.03 & 99.04 & 99.03 & 37.1 \\
\cmidrule{2-9}
 & \multirow{6}{*}{\makecell[l]{LBP +\\Poisson}}
   & 25 & n & 83.64 & 83.58 & 85.15 & 83.40 & 59.6 \\
 & & 25 & b & 83.60 & 83.55 & 85.11 & 83.36 & 59.7 \\
 & & 25 & m & 83.50 & 83.46 & 85.07 & 83.26 & 59.6 \\
 & & 5  & n & 83.63 & 83.63 & 85.24 & 83.41 & 35.6 \\
 & & 5  & b & \textbf{83.76} & \textbf{83.73} & 85.29 & 83.53 & 35.6 \\
 & & 5  & m & 83.66 & 83.62 & 85.15 & 83.42 & 35.6 \\
\cmidrule{2-9}
 & \multirow{6}{*}{\makecell[l]{LBP+CLBP\\+ Poisson}}
   & 25 & n & 81.36 & 80.99 & 83.81 & 80.96 & 81.3 \\
 & & 25 & b & 81.48 & 81.16 & 83.93 & 81.09 & 81.3 \\
 & & 25 & m & 81.50 & 81.18 & 83.92 & 81.12 & 81.2 \\
 & & 5  & n & 81.69 & 81.37 & 84.10 & 81.30 & 57.3 \\
 & & 5  & b & \textbf{81.92} & \textbf{81.63} & 84.20 & 81.55 & 57.2 \\
 & & 5  & m & 81.57 & 81.25 & 83.97 & 81.18 & 57.1 \\
\cmidrule{2-9}
 & \multirow{2}{*}{\makecell[l]{Poisson\\Only}}
   & 25 & n & 86.29 & 85.90 & 86.72 & 85.96 & 98.6 \\
 & & 25 & b & 86.23 & 85.82 & 86.65 & 85.89 & 98.7 \\
 & & 25 & m & 86.20 & 85.79 & 86.64 & 85.85 & 98.7 \\
 & & 5  & n & 86.37 & 85.98 & 86.77 & 86.03 & 74.8 \\
 & & 5  & b & \textbf{86.46} & \textbf{86.07} & 86.89 & 86.12 & 74.9 \\
 & & 5  & m & 86.38 & 85.99 & 86.81 & 86.05 & 74.9 \\
\midrule

\multirow{12}{*}{FMNIST}
 & \multirow{6}{*}{\makecell[l]{CNN +\\Poisson}}
   & 25 & n & 90.75 & 88.84 & 88.94 & 90.75 & 60.9 \\
 & & 25 & b & \textbf{90.76} & \textbf{88.81} & 88.92 & 90.76 & 60.9 \\
 & & 25 & m & 90.74 & 88.84 & 88.95 & 90.74 & 61.0 \\
 & & 5  & n & 89.08 & 88.92 & 89.02 & 89.08 & 36.9 \\
 & & 5  & b & 89.02 & 88.85 & 88.94 & 89.02 & 36.9 \\
 & & 5  & m & 89.05 & 88.87 & 88.95 & 89.05 & 36.8 \\
\cmidrule{2-9}
 & \multirow{6}{*}{\makecell[l]{LBP +\\Poisson}}
   & 25 & n & 65.20 & 61.48 & 67.96 & 65.13 & 59.6 \\
 & & 25 & b & 65.29 & 61.56 & 67.92 & 65.22 & 59.7 \\
 & & 25 & m & 65.00 & 61.28 & 67.81 & 64.95 & 59.8 \\
 & & 5  & n & 65.06 & 61.40 & 67.86 & 64.99 & 35.6 \\
 & & 5  & b & 65.08 & 61.44 & 67.79 & 65.02 & 35.6 \\
 & & 5  & m & \textbf{65.33} & \textbf{61.60} & 68.09 & 65.27 & 35.6 \\
\cmidrule{2-9}
 & \multirow{6}{*}{\makecell[l]{LBP+CLBP\\+ Poisson}}
   & 25 & n & 67.90 & 68.47 & 72.42 & 67.89 & 80.8 \\
 & & 25 & b & 67.91 & 68.48 & 72.19 & 67.91 & 80.8 \\
 & & 25 & m & \textbf{71.35} & \textbf{68.62} & 72.41 & 71.24 & 80.8 \\
 & & 5  & n & 71.25 & 68.51 & 72.19 & 71.25 & 56.8 \\
 & & 5  & b & 71.27 & 68.49 & 72.41 & 71.19 & 56.8 \\
 & & 5  & m & 71.20 & 68.40 & 72.24 & 71.20 & 80.7 \\
\cmidrule{2-9}
 & \multirow{6}{*}{\makecell[l]{Poisson\\Only}}
   & 25 & n & 76.44 & 75.19 & 76.19 & 76.44 & 98.9 \\
 & & 25 & b & \textbf{76.62} & \textbf{75.11} & 76.06 & 76.62 & 98.9 \\
 & & 25 & m & 76.57 & 75.29 & 76.55 & 76.57 & 98.9 \\
 & & 5  & n & 76.59 & 75.37 & 76.23 & 76.44 & 75.0 \\
 & & 5  & b & 76.56 & 75.36 & 76.34 & 76.56 & 75.0 \\
 & & 5  & m & 76.56 & 75.37 & 76.55 & 76.56 & 75.0 \\
\midrule

\multirow{12}{*}{CIFAR-10}
 & \multirow{6}{*}{\makecell[l]{CNN +\\Poisson}}
   & 25 & n & 70.39 & 68.68 & 70.79 & 70.39 & 50.1 \\
 & & 25 & b & 70.47 & 68.80 & 70.66 & 70.47 & 50.3 \\
 & & 25 & m & \textbf{70.53} & \textbf{68.84} & 70.88 & 70.53 & 50.4 \\
 & & 5  & n & 70.33 & 68.60 & 70.68 & 70.33 & 30.1 \\
 & & 5  & b & 70.47 & 68.80 & 70.66 & 70.47 & 30.1 \\
 & & 5  & m & 70.34 & 68.62 & 70.66 & 70.34 & 30.2 \\
\cmidrule{2-9}
 & \multirow{6}{*}{\makecell[l]{LBP +\\Poisson}}
   & 25 & n & 37.00 & 33.24 & 34.07 & 37.00 & 55.7 \\
 & & 25 & b & 36.98 & 33.24 & 33.97 & 36.98 & 55.6 \\
 & & 25 & m & 36.91 & 33.22 & 33.94 & 36.91 & 55.6 \\
 & & 5  & n & \textbf{37.30} & \textbf{33.62} & 34.25 & 37.30 & 35.5 \\
 & & 5  & b & 37.26 & 33.51 & 34.18 & 37.26 & 35.5 \\
 & & 5  & m & 37.05 & 33.40 & 34.06 & 37.05 & 35.5 \\
\cmidrule{2-9}
 & & 25 & n & 39.38 & 36.07 & 36.74 & 39.38 & 72.1 \\
 & & 25 & b & 39.46 & 36.12 & 36.81 & 39.46 & 72.1 \\
 & \multirow{3}{*}{\makecell[l]{LBP+CLBP\\+ Poisson}}
   & 25 & m & 39.46 & 36.13 & 36.86 & 39.46 & 72.1 \\
 & & 5  & n & 39.50 & 36.16 & 36.86 & 39.50 & 52.8 \\
 & & 5  & b & \textbf{39.64} & \textbf{36.30} & 36.99 & 39.64 & 52.6 \\
 & & 5  & m & 39.46 & 36.14 & 36.82 & 39.46 & 52.8 \\
\cmidrule{2-9}
 & \multirow{6}{*}{\makecell[l]{Poisson\\Only}}
   & 25 & n & 28.71 & 25.59 & 33.34 & 28.71 & 374.1 \\
 & & 25 & b & 29.82 & 26.78 & 34.19 & 29.82 & 440.8 \\
 & & 25 & m & 29.87 & 26.94 & 34.44 & 29.87 & 304.3 \\
 & & 5  & n & 26.78 & 23.53 & 32.19 & 26.78 & 357.2 \\
 & & 5  & b & \textbf{29.90} & \textbf{26.88} & 34.37 & 29.90 & 360.2 \\
 & & 5  & m & 29.68 & 26.88 & 33.56 & 29.68 & 358.9 \\
\end{longtable}

\subsection{STDP: Benchmark Datasets (Full Sweep)}
\label{app:stdp_bench}

\begin{longtable}{@{}llll rrrr r@{}}
\caption{%
  STDP full hyperparameter sweep: MNIST, FMNIST, CIFAR-10.
  $\tau$: time constant (ms). RM: reward mode (n=none, b=binary, m=margin).%
}
\label{tab:app_stdp_bench} \\
\toprule
\textbf{Dataset} & \textbf{Encoding} & $\tau$ & \textbf{RM}
  & \textbf{Acc.\%} & \textbf{F1\%} & \textbf{Prec.\%} & \textbf{Rec.\%}
  & \textbf{$t$/ep} \\
\midrule
\endfirsthead
\multicolumn{9}{l}{\small\itshape (continued from previous page)} \\
\toprule
\textbf{Dataset} & \textbf{Encoding} & $\tau$ & \textbf{RM}
  & \textbf{Acc.\%} & \textbf{F1\%} & \textbf{Prec.\%} & \textbf{Rec.\%}
  & \textbf{$t$/ep} \\
\midrule
\endhead
\midrule
\multicolumn{9}{r}{\small\itshape (continued on next page)} \\
\endfoot
\bottomrule
\endlastfoot

& \multirow{6}{*}{\makecell[l]{CNN +\\Poisson}}
   & 2.0  & n & 38.03 & 30.42 & 48.42 & 38.70 & 51.3 \\
 & & 2.0  & b & 48.40 & 43.58 & 66.02 & 49.23 & 51.4 \\
 & & 2.0  & m & \textbf{77.66} & \textbf{76.00} & 88.60 & 79.04 & 52.3 \\
 & & 10.0 & n & 27.48 & 17.17 & 16.87 & 26.19 & 51.1 \\
 & & 10.0 & b & 66.13 & 59.75 & 64.96 & 67.31 & 82.0 \\
 & & 10.0 & m & 37.09 & 28.87 & 51.01 & 37.87 & 102.4 \\
\cmidrule{2-9}
\multirow[c]{13}{*}{MNIST}
 & \multirow{6}{*}{\makecell[l]{LBP +\\Poisson}}
   & 2.0  & n & 10.26 &  2.81 & 10.79 & 10.53 & 52.6 \\
 & & 2.0  & b & 50.61 & 46.38 & 75.50 & 49.95 & 52.9 \\
 & & 2.0  & m & \textbf{78.91} & \textbf{78.39} & 82.25 & 78.74 & 52.3 \\
 & & 10.0 & n & 13.57 &  5.69 &  8.41 & 13.86 & 52.6 \\
 & & 10.0 & b & 48.96 & 44.36 & 74.13 & 48.21 & 54.2 \\
 & & 10.0 & m & 59.79 & 49.58 & 75.24 & 58.66 & 54.4 \\
\cmidrule{2-9}
 & \multirow{6}{*}{\makecell[l]{LBP+CLBP\\+ Poisson}}
   & 2.0  & n & 16.64 & 12.12 & 14.46 & 16.65 & 98.7 \\
 & & 2.0  & b & 50.75 & 47.12 & 63.55 & 50.39 & 99.8 \\
 & & 2.0  & m & \textbf{59.66} & \textbf{51.24} & 57.88 & 54.20 & 101.0 \\
 & & 10.0 & n &  9.58 &  1.75 &  0.96 & 10.00 & 100.1 \\
 & & 10.0 & b & 58.82 & 51.03 & 66.62 & 58.66 & 100.5 \\
 & & 10.0 & m & 54.50 & 44.46 & 57.88 & 54.20 & 99.6 \\
\cmidrule{2-9}
 & \multirow{6}{*}{\makecell[l]{Poisson\\Only}}
   & 2.0  & n & 12.99 &  7.87 & 26.99 & 13.77 & 138.7 \\
 & & 2.0  & b & 43.87 & 36.18 & 59.00 & 43.20 & 140.3 \\
 & & 2.0  & m & 53.98 & 48.28 & 66.32 & 53.82 & 139.7 \\
 & & 10.0 & n & 10.02 &  1.64 &  0.89 & 10.00 & 137.2 \\
 & & 10.0 & b & \textbf{62.80} & \textbf{57.79} & 64.19 & 62.21 & 139.0 \\
 & & 10.0 & m & 34.89 & 23.31 & 20.36 & 33.82 & 138.6 \\
\midrule

\multirow{12}{*}{FMNIST}
 & \multirow{6}{*}{\makecell[l]{CNN +\\Poisson}}
   & 2.0  & n & 48.74 & 41.61 & 52.74 & 48.74 & 52.0 \\
 & & 2.0  & b & 70.60 & 72.21 & 84.46 & 70.60 & 51.9 \\
 & & 2.0  & m & \textbf{74.25} & \textbf{74.11} & 88.03 & 74.25 & 52.0 \\
 & & 10.0 & n & 44.98 & 36.30 & 44.34 & 44.98 & 50.9 \\
 & & 10.0 & b & 28.18 & 25.27 & 55.43 & 28.18 & 51.1 \\
 & & 10.0 & m & 73.86 & 73.86 & 81.29 & 73.86 & 51.9 \\
\cmidrule{2-9}
 & \multirow{6}{*}{\makecell[l]{LBP +\\Poisson}}
   & 2.0  & n & 42.78 & 31.76 & 44.81 & 42.78 & 53.1 \\
 & & 2.0  & b & 62.63 & 56.24 & 65.47 & 62.63 & 52.7 \\
 & & 2.0  & m & \textbf{64.89} & \textbf{61.12} & 67.99 & 64.89 & 52.1 \\
 & & 10.0 & n & 30.71 & 20.09 & 23.13 & 30.71 & 53.3 \\
 & & 10.0 & b & 59.63 & 54.11 & 64.93 & 59.63 & 54.3 \\
 & & 10.0 & m & 63.71 & 59.58 & 67.37 & 63.71 & 53.8 \\
\cmidrule{2-9}
 & \multirow{6}{*}{\makecell[l]{LBP+CLBP\\+ Poisson}}
   & 2.0  & n & 20.00 & 11.60 & 22.31 & 20.00 & 98.7 \\
 & & 2.0  & b & 57.74 & 48.68 & 45.42 & 57.74 & 99.3 \\
 & & 2.0  & m & \textbf{69.66} & \textbf{66.01} & 74.68 & 69.66 & 98.9 \\
 & & 10.0 & n & 30.52 & 20.09 & 23.14 & 30.52 & 98.4 \\
 & & 10.0 & b & 60.33 & 51.32 & 53.22 & 60.33 & 99.4 \\
 & & 10.0 & m & 66.33 & 62.27 & 69.87 & 66.33 & 98.7 \\
\cmidrule{2-9}
 & \multirow{3}{*}{\makecell[l]{Poisson\\Only}}
   & 2.0  & n & 48.39 & 44.28 & 59.39 & 48.39 & 140.2 \\
 & & 2.0  & b & 47.55 & 39.55 & 62.93 & 47.55 & 140.3 \\
 & & 2.0  & m & \textbf{53.33} & \textbf{47.19} & 74.60 & 53.33 & 139.6 \\
 & & 10.0 & n & 37.53 & 25.43 & 23.13 & 37.53 & 138.5 \\
 & & 10.0 & b & 47.55 & 39.55 & 62.93 & 47.55 & 138.5 \\
 & & 10.0 & m & 49.07 & 39.95 & 54.10 & 49.07 & 140.0 \\
\midrule
\multirow{12}{*}{CIFAR-10}
 & \multirow{6}{*}{\makecell[l]{CNN +\\Poisson}}
   & 2.0  & n & 47.44 & 41.99 & 44.13 & 47.44 & 43.5 \\
 & & 2.0  & b & 64.72 & 62.70 & 68.04 & 64.72 & 43.7 \\
 & & 2.0  & m & \textbf{68.65} & \textbf{67.20} & 69.28 & 68.65 & 44.0 \\
 & & 10.0 & n & 27.64 & 16.87 & 38.60 & 27.64 & 42.6 \\
 & & 10.0 & b & 60.60 & 57.62 & 66.51 & 60.60 & 43.4 \\
 & & 10.0 & m & 61.90 & 59.48 & 65.94 & 61.90 & 43.5 \\
\cmidrule{2-9}
 & \multirow{6}{*}{\makecell[l]{LBP +\\Poisson}}
   & 2.0  & n & 10.00 &  1.82 &  1.00 & 10.00 & 56.3 \\
 & & 2.0  & b & 36.92 & 33.54 & 34.14 & 36.92 & 57.2 \\
 & & 2.0  & m & 37.27 & 34.09 & 34.97 & 37.27 & 57.4 \\
 & & 10.0 & n & 10.00 &  1.82 &  1.00 & 10.00 & 56.9 \\
 & & 10.0 & b & 37.73 & 34.72 & 43.09 & 37.73 & 57.8 \\
 & & 10.0 & m & \textbf{38.73} & \textbf{35.30} & 41.06 & 38.73 & 56.8 \\
\cmidrule{2-9}
 & \multirow{6}{*}{\makecell[l]{LBP+CLBP\\+ Poisson}}
   & 2.0  & n & 10.14 &  2.14 &  1.03 & 10.14 & 95.3 \\
 & & 2.0  & b & 41.88 & 38.66 & 43.92 & 41.88 & 95.0 \\
 & & 2.0  & m & 40.40 & 37.26 & 42.75 & 40.40 & 97.6 \\
 & & 10.0 & n & 10.90 &  3.65 &  4.66 & 10.90 & 96.3 \\
 & & 10.0 & b & 41.63 & 38.27 & 46.67 & 41.63 & 95.9 \\
 & & 10.0 & m & \textbf{41.96} & \textbf{38.63} & 45.10 & 41.96 & 95.2 \\
\cmidrule{2-9}
 & \multirow{6}{*}{\makecell[l]{Poisson\\Only}}
   & 2.0  & n & 10.14 &  2.14 &  1.03 & 10.14 & 602.9 \\
 & & 2.0  & b & 22.80 & 17.74 & 40.11 & 22.80 & 598.5 \\
 & & 2.0  & m & \textbf{28.38} & \textbf{24.74} & 35.08 & 28.38 & 602.7 \\
 & & 10.0 & n &  9.85 &  3.03 &  4.66 &  9.85 & 597.0 \\
 & & 10.0 & b & 11.55 &  3.85 &  2.40 & 11.55 & 611.9 \\
 & & 10.0 & m & 18.27 &  9.43 & 23.79 & 18.27 & 606.9 \\
\end{longtable}

\subsection{SADP: Medical Imaging Datasets (Full Sweep)}
\label{app:sadp_med}

\begin{longtable}{@{}llll rrrr r@{}}
\caption{%
  SADP full hyperparameter sweep: COLON, LUNG, TUMOUR.%
}
\label{tab:app_sadp_med} \\
\toprule
\textbf{Dataset} & \textbf{Encoding} & $K$ & \textbf{RM}
  & \textbf{Acc.\%} & \textbf{F1\%} & \textbf{Prec.\%} & \textbf{Rec.\%}
  & \textbf{$t$/ep} \\
\midrule
\endfirsthead
\multicolumn{9}{l}{\small\itshape (continued from previous page)} \\
\toprule
\textbf{Dataset} & \textbf{Encoding} & $K$ & \textbf{RM}
  & \textbf{Acc.\%} & \textbf{F1\%} & \textbf{Prec.\%} & \textbf{Rec.\%}
  & \textbf{$t$/ep} \\
\midrule
\endhead
\midrule
\multicolumn{9}{r}{\small\itshape (continued on next page)} \\
\endfoot
\bottomrule
\endlastfoot

\multirow{12}{*}{COLON}
 & \multirow{6}{*}{\makecell[l]{CNN +\\Poisson}}
   & 25 & n & 98.75 & 98.75 & 98.75 & 98.75 & 8.4 \\
 & & 25 & b & 98.75 & 98.75 & 98.75 & 98.75 & 8.3 \\
 & & 25 & m & 99.50 & 99.50 & 99.50 & 99.50 & 8.3 \\
 & & 5  & n & 99.50 & 99.50 & 99.50 & 99.50 & 5.0 \\
 & & 5  & b & 99.50 & 99.50 & 99.50 & 99.50 & 4.9 \\
 & & 5  & m & \textbf{99.60} & \textbf{99.60} & 99.60 & 99.60 & 4.9 \\
\cmidrule{2-9}
 & \multirow{6}{*}{\makecell[l]{LBP +\\Poisson}}
   & 25 & n & 79.90 & 79.87 & 80.43 & 79.90 & 18.2 \\
 & & 25 & b & 80.85 & 80.81 & 81.26 & 80.85 & 15.5 \\
 & & 25 & m & 80.95 & 80.91 & 81.38 & 80.95 & 9.2 \\
 & & 5  & n & 81.30 & 81.26 & 81.54 & 81.30 & 11.3 \\
 & & 5  & b & \textbf{81.55} & \textbf{81.52} & 81.73 & 81.55 & 11.3 \\
 & & 5  & m & 80.80 & 80.76 & 81.19 & 80.80 & 11.3 \\
\cmidrule{2-9}
 & \multirow{6}{*}{\makecell[l]{LBP+CLBP\\+ Poisson}}
   & 25 & n & \textbf{76.55} & \textbf{76.34} & 77.53 & 76.55 & 21.5 \\
 & & 25 & b & 76.50 & 76.29 & 77.49 & 76.50 & 21.5 \\
 & & 25 & m & 74.90 & 74.71 & 75.65 & 74.90 & 21.5 \\
 & & 5  & n & 75.75 & 75.53 & 76.62 & 75.75 & 15.0 \\
 & & 5  & b & 75.80 & 75.57 & 76.68 & 75.80 & 15.0 \\
 & & 5  & m & 76.35 & 76.13 & 77.15 & 76.35 & 15.0 \\
\cmidrule{2-9}
 & \multirow{6}{*}{\makecell[l]{Poisson\\Only}}
   & 25 & n & 50.00 & 33.33 & 25.00 & 50.00 & 207.3 \\
 & & 25 & b & 52.36 & 44.05 & 38.52 & 52.36 & 293.5 \\
 & & 25 & m & 52.71 & 37.11 & 34.10 & 52.71 & 341.1 \\
 & & 5  & n & 50.00 & 33.33 & 25.00 & 50.00 & 181.1 \\
 & & 5  & b & \textbf{56.50} & \textbf{50.96} & 44.17 & 56.50 & 172.7 \\
 & & 5  & m & 51.15 & 37.11 & 33.88 & 51.15 & 168.1 \\
\midrule

 & \multirow{6}{*}{\makecell[l]{CNN +\\Poisson}}
   & 25 & n & 96.73 & 96.72 & 96.74 & 96.73 & 22.1 \\
 & & 25 & b & 96.73 & 96.72 & 96.74 & 96.73 & 20.6 \\
 & & 25 & m & 97.20 & 97.19 & 97.21 & 97.20 & 19.9 \\
 & & 5  & n & 97.27 & 97.26 & 97.27 & 97.27 & 7.7 \\
 & & 5  & b & 97.20 & 97.19 & 97.20 & 97.20 & 7.6 \\
 & & 5  & m & \textbf{97.30} & \textbf{97.29} & 97.30 & 97.30 & 7.7 \\
\cmidrule{2-9}
\multirow{12}{*}{LUNG}
 & \multirow{6}{*}{\makecell[l]{LBP +\\Poisson}}
   & 25 & n & 83.10 & 82.76 & 83.48 & 83.10 & 14.9 \\
 & & 25 & b & 83.70 & 83.38 & 84.73 & 83.70 & 14.9 \\
 & & 25 & m & 83.57 & 83.24 & 84.58 & 83.57 & 14.8 \\
 & & 5  & n & \textbf{83.80} & \textbf{83.48} & 84.83 & 83.80 & 9.3 \\
 & & 5  & b & 83.43 & 83.10 & 84.32 & 83.43 & 9.4 \\
 & & 5  & m & 83.53 & 83.19 & 84.52 & 83.53 & 9.3 \\
\cmidrule{2-9}
 & \multirow{6}{*}{\makecell[l]{LBP+CLBP\\+ Poisson}}
   & 25 & n & 79.87 & 78.31 & 80.19 & 79.87 & 19.4 \\
 & & 25 & b & 79.93 & 78.40 & 80.28 & 79.93 & 19.2 \\
 & & 25 & m & \textbf{80.07} & \textbf{78.54} & 80.45 & 80.07 & 19.3 \\
 & & 5  & n & 79.73 & 78.17 & 80.06 & 79.73 & 13.9 \\
 & & 5  & b & 79.80 & 78.25 & 80.13 & 79.80 & 14.0 \\
 & & 5  & m & 79.87 & 78.32 & 80.23 & 79.87 & 13.8 \\
\cmidrule{2-9}
 & \multirow{6}{*}{\makecell[l]{Poisson\\Only}}
   & 25 & n & 73.20 & 70.67 & 74.93 & 73.20 & 256.1 \\
 & & 25 & b & \textbf{73.33} & \textbf{70.43} & 74.94 & 73.33 & 254.8 \\
 & & 25 & m & 72.90 & 70.00 & 74.94 & 72.90 & 273.6 \\
 & & 5  & n & 69.33 & 63.25 & 74.36 & 69.33 & 251.6 \\
 & & 5  & b & 61.35 & 47.88 & 39.86 & 61.35 & 314.5 \\
 & & 5  & m & 66.50 & 59.42 & 71.70 & 66.50 & 256.4 \\
\midrule

 & \multirow{3}{*}{\makecell[l]{CNN +\\Poisson}}
   & 25 & n & 84.32 & 84.30 & 83.99 & 84.67 & 19.1 \\
 & & 25 & b & 84.32 & 84.33 & 84.00 & 84.68 & 19.1 \\
 & & 25 & m & 84.34 & 84.31 & 83.99 & 84.68 & 10.0 \\
 & & 5  & n & \textbf{84.60} & \textbf{84.59} & 84.00 & 85.69 & 11.1 \\
 & & 5  & b & 84.32 & 84.33 & 84.00 & 84.68 & 13.8 \\
 & & 5  & m & 84.34 & 84.31 & 83.99 & 84.68 & 11.0 \\
\cmidrule{2-9}
\multirow{12}{*}{TUMOUR}
 & \multirow{6}{*}{\makecell[l]{LBP +\\Poisson}}
   & 25 & n & \textbf{51.65} & \textbf{41.59} & 38.23 & 51.65 & 22.6 \\
 & & 25 & b & 51.37 & 41.39 & 38.18 & 51.37 & 22.5 \\
 & & 25 & m & 51.56 & 41.53 & 38.17 & 51.56 & 22.6 \\
 & & 5  & n & 51.37 & 41.39 & 38.17 & 51.37 & 13.8 \\
 & & 5  & b & 51.60 & 41.53 & 38.21 & 51.60 & 13.8 \\
 & & 5  & m & 51.37 & 41.37 & 38.12 & 51.37 & 13.8 \\
\cmidrule{2-9}
 & \multirow{6}{*}{\makecell[l]{LBP+CLBP\\+ Poisson}}
   & 25 & n & 53.19 & 45.65 & 41.16 & 53.19 & 29.7 \\
 & & 25 & b & 53.27 & 45.63 & 41.14 & 53.27 & 29.6 \\
 & & 25 & m & 53.38 & 45.95 & 41.25 & 53.38 & 29.5 \\
 & & 5  & n & 53.37 & 45.67 & 41.17 & 53.37 & 20.7 \\
 & & 5  & b & \textbf{53.42} & \textbf{45.93} & 41.26 & 53.42 & 20.5 \\
 & & 5  & m & 53.27 & 45.74 & 41.27 & 53.27 & 20.7 \\
\cmidrule{2-9}
 & \multirow{6}{*}{\makecell[l]{Poisson\\Only}}
   & 25 & n & 50.82 & 40.35 & 38.01 & 50.82 & 384.6 \\
 & & 25 & b & \textbf{51.58} & \textbf{40.97} & 38.08 & 51.58 & 377.3 \\
 & & 25 & m & 51.16 & 40.62 & 38.03 & 51.16 & 372.1 \\
 & & 5  & n & 50.71 & 40.35 & 38.02 & 50.71 & 461.3 \\
 & & 5  & b & 50.99 & 40.60 & 38.04 & 50.99 & 372.1 \\
 & & 5  & m & 51.16 & 40.62 & 38.04 & 51.16 & 372.1 \\
\end{longtable}

\subsection{STDP: Medical Imaging Datasets (Full Sweep)}
\label{app:stdp_med}

\begin{longtable}{@{}llll rrrr r@{}}
\caption{%
  STDP full hyperparameter sweep: COLON, LUNG, TUMOUR.%
}
\label{tab:app_stdp_med} \\
\toprule
\textbf{Dataset} & \textbf{Encoding} & $\tau$ & \textbf{RM}
  & \textbf{Acc.\%} & \textbf{F1\%} & \textbf{Prec.\%} & \textbf{Rec.\%}
  & \textbf{$t$/ep} \\
\midrule
\endfirsthead
\multicolumn{9}{l}{\small\itshape (continued from previous page)} \\
\toprule
\textbf{Dataset} & \textbf{Encoding} & $\tau$ & \textbf{RM}
  & \textbf{Acc.\%} & \textbf{F1\%} & \textbf{Prec.\%} & \textbf{Rec.\%}
  & \textbf{$t$/ep} \\
\midrule
\endhead
\midrule
\multicolumn{9}{r}{\small\itshape (continued on next page)} \\
\endfoot
\bottomrule
\endlastfoot

\multirow{12}{*}{COLON}
 & \multirow{6}{*}{\makecell[l]{CNN +\\Poisson}}
   & 2.0  & n & 99.70 & 99.70 & 99.70 & 99.70 & 6.9 \\
 & & 2.0  & b & 99.75 & 99.75 & 99.75 & 99.75 & 6.6 \\
 & & 2.0  & m & \textbf{99.80} & \textbf{99.80} & 99.80 & 99.80 & 6.8 \\
 & & 10.0 & n & 50.00 & 33.33 & 25.00 & 50.00 & 6.9 \\
 & & 10.0 & b & 50.00 & 33.33 & 25.00 & 50.00 & 7.1 \\
 & & 10.0 & m & 50.00 & 33.33 & 25.00 & 50.00 & 7.3 \\
\cmidrule{2-9}
 & \multirow{6}{*}{\makecell[l]{LBP +\\Poisson}}
   & 2.0  & n & 69.35 & 68.21 & 71.05 & 69.35 & 8.6 \\
 & & 2.0  & b & 79.05 & 79.04 & 79.17 & 79.05 & 8.7 \\
 & & 2.0  & m & \textbf{79.90} & \textbf{79.85} & 80.10 & 79.90 & 8.8 \\
 & & 10.0 & n & 64.55 & 63.57 & 66.87 & 64.55 & 8.6 \\
 & & 10.0 & b & 72.85 & 72.77 & 73.23 & 72.85 & 9.2 \\
 & & 10.0 & m & 74.00 & 73.95 & 74.48 & 74.00 & 9.4 \\
\cmidrule{2-9}
 & \multirow{6}{*}{\makecell[l]{LBP+CLBP\\+ Poisson}}
   & 2.0  & n & 64.70 & 64.69 & 64.75 & 64.70 & 14.8 \\
 & & 2.0  & b & 72.55 & 72.45 & 73.13 & 72.55 & 14.9 \\
 & & 2.0  & m & \textbf{75.80} & \textbf{75.63} & 76.38 & 75.80 & 14.8 \\
 & & 10.0 & n & 50.00 & 33.33 & 25.00 & 50.00 & 14.8 \\
 & & 10.0 & b & 65.50 & 65.24 & 66.69 & 65.50 & 15.9 \\
 & & 10.0 & m & 70.65 & 70.45 & 71.68 & 70.65 & 16.3 \\
\cmidrule{2-9}
 & & 2.0  & n & 50.00 & 33.33 & 25.00 & 50.00 & 555.4 \\
 & \multirow{6}{*}{\makecell[l]{Poisson\\Only}}
   & 2.0  & b & 50.00 & 33.33 & 25.00 & 50.00 & 679.8 \\
 & & 2.0  & m & 50.00 & 33.33 & 25.00 & 50.00 & 444.6 \\
 & & 10.0 & n & \textbf{50.00} & \textbf{33.33} & 25.00 & 50.00 & 483.4 \\
 & & 10.0 & b & 50.00 & 33.33 & 25.00 & 50.00 & 440.7 \\
 & & 10.0 & m & 50.00 & 33.33 & 25.00 & 50.00 & 432.9 \\
\midrule

\multirow{12}{*}{LUNG}
 & \multirow{6}{*}{\makecell[l]{CNN +\\Poisson}}
   & 2.0  & n & 96.77 & 96.76 & 96.77 & 96.77 & 9.3 \\
 & & 2.0  & b & \textbf{96.93} & \textbf{96.93} & 97.00 & 96.93 & 9.2 \\
 & & 2.0  & m & 96.93 & 96.93 & 97.00 & 96.93 & 9.2 \\
 & & 10.0 & n & 86.97 & 86.23 & 87.90 & 86.97 & 9.7 \\
 & & 10.0 & b & 89.83 & 89.42 & 90.20 & 89.83 & 9.7 \\
 & & 10.0 & m & 89.43 & 88.95 & 90.07 & 89.43 & 10.1 \\
\cmidrule{2-9}
 & \multirow{6}{*}{\makecell[l]{LBP +\\Poisson}}
   & 2.0  & n & 70.90 & 67.01 & 72.06 & 70.90 & 12.0 \\
 & & 2.0  & b & 76.20 & 73.92 & 77.52 & 76.20 & 12.4 \\
 & & 2.0  & m & \textbf{82.80} & \textbf{82.43} & 84.54 & 82.80 & 12.0 \\
 & & 10.0 & n & 67.53 & 63.08 & 69.62 & 67.53 & 12.2 \\
 & & 10.0 & b & 72.20 & 69.05 & 74.22 & 72.20 & 11.9 \\
 & & 10.0 & m & 81.70 & 81.23 & 83.25 & 81.70 & 12.1 \\
\cmidrule{2-9}
 & \multirow{6}{*}{\makecell[l]{LBP+CLBP\\+ Poisson}}
   & 2.0  & n & 55.33 & 44.62 & 60.20 & 55.33 & 20.8 \\
 & & 2.0  & b & 64.33 & 51.75 & 64.20 & 64.33 & 20.8 \\
 & & 2.0  & m & \textbf{73.67} & \textbf{68.98} & 76.33 & 73.67 & 21.4 \\
 & & 10.0 & n & 48.60 & 39.42 & 56.30 & 48.60 & 21.0 \\
 & & 10.0 & b & 61.87 & 51.09 & 62.13 & 61.87 & 21.4 \\
 & & 10.0 & m & 68.17 & 62.39 & 70.68 & 68.17 & 21.2 \\
\cmidrule{2-9}
 & \multirow{6}{*}{\makecell[l]{Poisson\\Only}}
   & 2.0  & n & 39.10 & 31.98 & 45.86 & 39.10 & 801.3 \\
 & & 2.0  & b & 54.87 & 52.41 & 61.26 & 54.87 & 685.8 \\
 & & 2.0  & m & \textbf{62.70} & \textbf{52.39} & 75.08 & 62.70 & 640.3 \\
 & & 10.0 & n & 41.43 & 33.00 & 43.76 & 41.43 & 734.0 \\
 & & 10.0 & b & 47.03 & 41.73 & 55.70 & 47.03 & 727.8 \\
 & & 10.0 & m & 46.73 & 38.92 & 57.40 & 46.73 & 724.3 \\
\midrule

\multirow{12}{*}{TUMOUR}
 & \multirow{6}{*}{\makecell[l]{CNN +\\Poisson}}
   & 2.0  & n & 57.48 & 57.31 & 55.87 & 58.83 & 13.2 \\
 & & 2.0  & b & 80.40 & 80.95 & 80.93 & 81.01 & 13.2 \\
 & & 2.0  & m & \textbf{83.61} & \textbf{83.74} & 83.39 & 84.44 & 13.3 \\
 & & 10.0 & n & 37.09 & 28.87 & 51.01 & 37.87 & 13.1 \\
 & & 10.0 & b & 68.35 & 70.12 & 70.55 & 69.82 & 13.5 \\
 & & 10.0 & m & 73.67 & 74.35 & 74.17 & 74.56 & 13.2 \\
\cmidrule{2-9}
 & \multirow{6}{*}{\makecell[l]{LBP +\\Poisson}}
   & 2.0  & n & 38.16 & 28.27 & 35.18 & 38.16 & 18.3 \\
 & & 2.0  & b & 51.30 & 41.28 & 38.07 & 51.30 & 18.3 \\
 & & 2.0  & m & 51.30 & 41.28 & 38.07 & 51.30 & 18.6 \\
 & & 10.0 & n & 47.74 & 36.73 & 37.88 & 47.74 & 18.9 \\
 & & 10.0 & b & \textbf{52.46} & \textbf{43.09} & 38.82 & 52.46 & 18.9 \\
 & & 10.0 & m & 51.02 & 41.50 & 38.10 & 51.02 & 18.5 \\
\cmidrule{2-9}
 & \multirow{6}{*}{\makecell[l]{LBP+CLBP\\+ Poisson}}
   & 2.0  & n & 38.74 & 28.97 & 36.07 & 38.74 & 31.0 \\
 & & 2.0  & b & 52.57 & 41.89 & 38.11 & 52.57 & 31.2 \\
 & & 2.0  & m & 53.45 & 46.06 & 41.37 & 53.45 & 31.1 \\
 & & 10.0 & n & 37.63 & 27.77 & 34.87 & 37.63 & 31.4 \\
 & & 10.0 & b & \textbf{55.66} & \textbf{50.78} & 43.17 & 55.66 & 31.4 \\
 & & 10.0 & m & 53.53 & 46.07 & 41.36 & 53.53 & 31.2 \\
\cmidrule{2-9}
 & \multirow{6}{*}{\makecell[l]{Poisson\\Only}}
   & 2.0  & n & 29.12 & 11.31 & 28.38 & 29.12 & 808.1 \\
 & & 2.0  & b & \textbf{52.57} & \textbf{41.89} & 38.11 & 52.57 & 1078.0 \\
 & & 2.0  & m & 50.52 & 40.32 & 38.02 & 50.52 & 931.8 \\
 & & 10.0 & n & 41.60 & 31.72 & 37.60 & 41.60 & 791.2 \\
 & & 10.0 & b & 51.27 & 40.96 & 38.07 & 51.27 & 788.9 \\
 & & 10.0 & m & 50.52 & 40.32 & 38.02 & 50.52 & 793.2 \\
\end{longtable}

\begin{table}[htbp]
\centering
\caption{%
  Full per-cell training speedup. Each row reports the best-accuracy
  configuration of each algorithm (the same configs as
  Tables~\ref{tab:benchmark} and~\ref{tab:medical}), its time per epoch
  ($t$/ep, s), and the speedup $s = t^{\text{STDP}}/t^{\text{SADP}}$ from
  Eq.~\eqref{eq:speedup}. RM: reward mode (n=none, b=binary, m=margin).
  Values $>1$ mean SADP is faster. For COLON Poisson-only, all six STDP
  configurations sit at chance (50.00\%); the $t$/ep shown is that of the
  fastest such configuration.%
}
\label{tab:speed_full}
\setlength{\tabcolsep}{5pt}
\renewcommand{\arraystretch}{1.05}
\begin{tabular}{@{}ll cc c cc c r@{}}
\toprule
& &
\multicolumn{3}{c}{\textbf{SADP}} &
\multicolumn{3}{c}{\textbf{STDP}} & \\
\cmidrule(lr){3-5}\cmidrule(lr){6-8}
\textbf{Dataset} & \textbf{Encoding} &
  $K$ & RM & $t$/ep &
  $\tau$ & RM & $t$/ep &
  \textbf{Speedup} \\
\midrule
\multirow{4}{*}{MNIST}
 & Poisson Only        & 5 & b &   74.9 & 10.0 & b &   139.0 & $1.86\times$ \\
 & LBP + Poisson       & 5 & b &   35.6 & 2.0 & m &    52.3 & $1.47\times$ \\
 & LBP+CLBP + Poisson  & 5 & b &   57.2 & 2.0 & m &   101.0 & $1.77\times$ \\
 & CNN + Poisson       & 5 & n &   36.9 & 2.0 & m &    52.3 & $1.42\times$ \\
\midrule
\multirow{4}{*}{FMNIST}
 & Poisson Only        & 25 & b &   98.9 & 2.0 & m &   139.6 & $1.41\times$ \\
 & LBP + Poisson       & 5 & m &   35.6 & 2.0 & m &    52.1 & $1.46\times$ \\
 & LBP+CLBP + Poisson  & 25 & m &   80.8 & 2.0 & m &    98.9 & $1.22\times$ \\
 & CNN + Poisson       & 25 & b &   60.9 & 2.0 & m &    52.0 & $0.85\times$ \\
\midrule
\multirow{4}{*}{CIFAR-10}
 & Poisson Only        & 5 & b &  360.2 & 2.0 & m &   602.7 & $1.67\times$ \\
 & LBP + Poisson       & 5 & n &   35.5 & 10.0 & m &    56.8 & $1.60\times$ \\
 & LBP+CLBP + Poisson  & 5 & b &   52.6 & 10.0 & m &    95.2 & $1.81\times$ \\
 & CNN + Poisson       & 25 & m &   50.4 & 2.0 & m &    44.0 & $0.87\times$ \\
\midrule
\multirow{4}{*}{COLON}
 & Poisson Only        & 5 & b &  172.7 & 10.0 & m &   432.9 & $2.51\times$ \\
 & LBP + Poisson       & 5 & b &   11.3 & 2.0 & m &     8.8 & $0.78\times$ \\
 & LBP+CLBP + Poisson  & 25 & n &   21.5 & 2.0 & m &    14.8 & $0.69\times$ \\
 & CNN + Poisson       & 5 & m &    4.9 & 2.0 & m &     6.8 & $1.39\times$ \\
\midrule
\multirow{4}{*}{LUNG}
 & Poisson Only        & 25 & b &  254.8 & 2.0 & m &   640.3 & $2.51\times$ \\
 & LBP + Poisson       & 5 & n &    9.3 & 2.0 & m &    12.0 & $1.29\times$ \\
 & LBP+CLBP + Poisson  & 25 & m &   19.3 & 2.0 & m &    21.4 & $1.11\times$ \\
 & CNN + Poisson       & 5 & m &    7.7 & 2.0 & b &     9.2 & $1.19\times$ \\
\midrule
\multirow{4}{*}{TUMOUR}
 & Poisson Only        & 25 & b &  377.3 & 2.0 & b &  1078.0 & $2.86\times$ \\
 & LBP + Poisson       & 25 & n &   22.6 & 10.0 & b &    18.9 & $0.84\times$ \\
 & LBP+CLBP + Poisson  & 5 & b &   20.5 & 10.0 & b &    31.4 & $1.53\times$ \\
 & CNN + Poisson       & 5 & n &   11.1 & 2.0 & m &    13.3 & $1.20\times$ \\
\midrule
\multicolumn{8}{r}{\textbf{Overall mean speedup}} & $\mathbf{1.47\times}$ \\
\bottomrule
\end{tabular}
\end{table}

\section{Device-Calibrated Supervised SADP: Hardware Compatibility Validation}
\label{app:device}

\paragraph{Motivation.}
A key claim of the Supervised SADP framework is its compatibility with
neuromorphic hardware.
To provide concrete evidence of this compatibility, we replaced the
linear $\kappa$-to-weight-update mapping with a kernel derived from
experimentally measured conductance dynamics of an iontronic memtransistor
device.
This experiment should be viewed as a \emph{device-inspired validation}
rather than a full hardware deployment: the goal is to demonstrate that
the SADP learning rule is inherently compatible with emerging memtransistor
technologies, without requiring any algorithmic modification.

\begin{figure}[!htbp]
    \centering
    \includegraphics[width=0.9\linewidth]{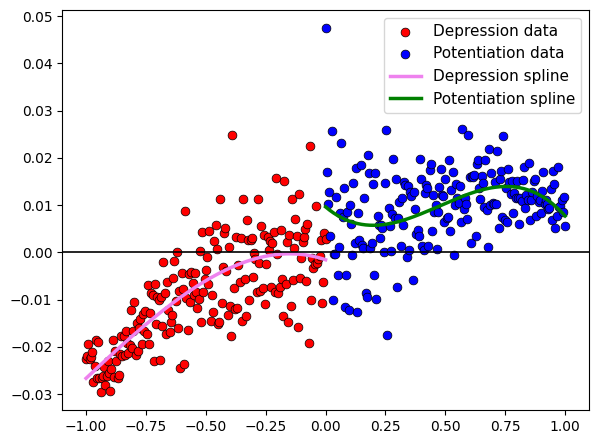}
    \caption{%
    Device kernel extracted from a physical iontronic memtransistor.
    Measured potentiation (blue) and depression (red) conductance samples,
    with the fitted smoothing splines (green and pink) that map the spike
    agreement $\kappa\in[-1,1]$ to the normalised conductance change
    $\Delta G/G_0$. Positive $\kappa$ (co-firing with the correct-class
    output) drives potentiation; negative $\kappa$ drives depression. This
    device response $f_{\text{device}}(\kappa)$ replaces the identity mapping
    $\kappa\mapsto\kappa$ in the standard Supervised SADP update.}
    \label{fig:device_plot}
\end{figure}

\paragraph{Device kernel construction.}
Potentiation and depression conductance trajectories were measured from a
physical memtransistor device and fitted using smooth spline interpolation
(smoothing parameters $s_\text{pot}=0.1$, $s_\text{dep}=0.01$).
The resulting spline functions map the spike agreement signal
$\kappa \in [-1,1]$ directly to the normalised conductance change
$\Delta G / G_0$ measured from the device:
positive $\kappa$ (hidden neuron co-fires with the correct-class output)
drives potentiation; negative $\kappa$ drives depression.
This mapping replaces the identity $\kappa \mapsto \kappa$ in the standard
SADP update with the physical device response
$\kappa \mapsto f_\text{device}(\kappa)$,
preserving all other aspects of the algorithm unchanged.
The measured conductance samples and the fitted spline kernel are shown in
Figure~\ref{fig:device_plot}.

\paragraph{Experimental setup.}
Device-calibrated SADP was evaluated on MNIST and Fashion-MNIST using
LBP+Poisson, LBP+CLBP+Poisson, and CNN+Poisson encodings with
$K{=}5$, \texttt{none} reward mode, and seed~42.
Table~\ref{tab:device} compares device-calibrated accuracy against the
standard SADP result for the same configuration.

\paragraph{Results.}
The device kernel produces accuracy within $1.5$ pp of the
standard linear-$\kappa$ result in all six configurations tested
(differences ranging from $+0.04$ to $+1.45$ pp), and is marginally
\emph{higher} in all six cases.
The largest differences occur for LBP+CLBP+Poisson ($+1.43$ pp on MNIST,
$+1.45$ pp on Fashion-MNIST), where the non-linear device response appears
to act as a mild regulariser, compressing extreme $\kappa$ values and
smoothing the weight update.
The CNN+Poisson results are nearly identical (within $0.1$ pp), reflecting
that the CNN features already provide strong discriminative structure that
is insensitive to the precise shape of the plasticity kernel.
These results demonstrate that Supervised SADP is directly compatible with
real memtransistor conductance profiles: the physical non-linearity and
asymmetry of the device do not degrade learning, and may in fact provide a
beneficial inductive bias for texture-based encodings.

\begin{table}[ht]
\centering
\caption{%
Device-calibrated Supervised SADP versus standard SADP.
Results use the potentiation/depression conductance spline extracted from
a physical iontronic memtransistor device as the weight-update kernel,
replacing the linear $\kappa \mapsto \kappa$ mapping.
Configuration: $K=5$, reward mode \texttt{none}, seed 42, $T=50$.
$\Delta$ = Device Acc. $-$ Standard Acc. (pp).}
\label{tab:device}

\small
\setlength{\tabcolsep}{4pt}
\renewcommand{\arraystretch}{1.1}

\begin{tabularx}{\linewidth}{@{}lXccc@{}}
\toprule
\textbf{Dataset} &
\textbf{Encoding} &
\textbf{Std.} &
\textbf{Device} &
$\mathbf{\Delta}$ \\
\midrule
\multirow{3}{*}{MNIST}
& LBP + Poisson      & 83.63 & 83.67 & +0.04 \\
& LBP+CLBP + Poisson & 81.69 & 83.12 & +1.43 \\
& CNN + Poisson      & 99.05 & 99.10 & +0.05 \\
\midrule
\multirow{3}{*}{FMNIST}
& LBP + Poisson      & 65.06 & 65.39 & +0.33 \\
& LBP+CLBP + Poisson & 71.25 & 72.70 & +1.45 \\
& CNN + Poisson      & 89.08 & 89.17 & +0.09 \\
\bottomrule
\end{tabularx}
\end{table}

\end{document}